%% file: main.tex
\documentclass[10.5pt,twocolumn,letterpaper]{article}

\usepackage[pagenumbers]{cvpr}
\input{headers}

\begin{document}

\title{\LARGE Demonstrating CavePI: Autonomous Exploration of \\Underwater Caves by Semantic Guidance
\vspace{-5mm}
}%
\author{Alankrit Gupta$^{\star 1}$, Adnan Abdullah$^{\star 1}$, Xianyao Li$^1$, Vaishnav Ramesh$^1$, \\ Ioannis Rekleitis$^2$, and Md Jahidul Islam$^1$\\
{\small$^1$RoboPI laboratory, Department of ECE, University of Florida, FL 32611, USA}\\
{\small$^2$Department of ME, University of Delaware, Newark, DE 19716, USA}\\
{
\thanks{$^\star$ These two authors have contributed equally.}
\thanks{Accepted for publication at the Robotics: Science and Systems (RSS), 2025, Los Angeles, CA, US.}
}
\vspace{-5mm}
}
\maketitle

\input{src/00_Abstract}

\input{src/01_Introduction}

\input{src/02_Background_Related}
\input{src/03_System_Design}
\input{src/04_Nav_Pipeline}
\input{src/05_System_Evaluation}

\input{src/06_Field_Trials}
\input{src/07_Limitation}

\input{src/08_Review_Discussion}

\input{src/09_Conclusion}

{\small
\bibliographystyle{plainnat}
\bibliography{references,robopi_pubs,old_cave_refs}
}

\end{document}

%% file: headers.tex
\usepackage{times}
\usepackage[numbers]{natbib}
\usepackage{multicol}
\usepackage[bookmarks=true]{hyperref}

\usepackage{amsmath,amssymb,amsfonts}
\usepackage{graphicx}
\usepackage{multirow}
\usepackage{soul}
\usepackage{siunitx}
\usepackage{physunits}
\usepackage{xspace,xcolor}
\usepackage{algorithm}
\usepackage{algorithmic}
\usepackage{array}

\usepackage{makecell}
\usepackage{subcaption}
\usepackage[size=small]{caption}
\usepackage{booktabs}
\makeatletter
\DeclareRobustCommand\onedot{\futurelet\@let@token\@onedot}
\def\@onedot{\ifx\@let@token.\else.\null\fi\xspace}
\def\eg{\emph{e.g}\onedot} 
\def\ie{\emph{i.e}\onedot}

\def\etal{\emph{et al}\onedot}
\makeatother

\usepackage{wrapfig}
\usepackage{makecell}
\long\def\invis#1{}

\let\labelindent\relax
\usepackage[shortlabels]{enumitem}

\newcommand\z{\phantom{0}}

\newcommand{\codecomment}[1]{\textcolor{gray}{\textit{//#1}}}

\usepackage{lipsum}

%% file: src/00_Abstract.tex
\begin{abstract}
Enabling autonomous robots to safely and efficiently navigate, explore, and map underwater caves is of significant importance to water resource management, hydrogeology,  archaeology, and marine robotics. In this work, we demonstrate the system design and algorithmic integration of a visual servoing framework for semantically guided autonomous underwater cave exploration. We present the hardware and edge-AI design considerations for deploying this framework on a novel AUV (Autonomous Underwater Vehicle) named CavePI. The guided navigation is driven by a computationally light yet robust deep visual perception module, delivering a rich semantic understanding of the environment. Subsequently, a robust control mechanism enables CavePI to track the semantic guides and navigate within complex cave structures. We evaluate the CavePI system through field experiments in natural underwater caves and spring-water sites, and further validate its ROS (Robot Operating System)-based digital twin in a simulation environment. Our results highlight how these integrated design choices facilitate reliable AUV navigation under feature-deprived, GPS-denied, and low-visibility conditions with overhead obstacles. The system design, code, and data are available at: \textbf{\url{https://github.com/uf-robopi/CavePI_AUV/}}.

\vspace{1.5mm}
\noindent
\textbf{Keywords.} Autonomous Underwater Vehicles (AUVs); Guided Navigation; Underwater Cave Exploration.
\end{abstract}

%% file: src/01_Introduction.tex
\section{Introduction}
Autonomous underwater cave exploration is an important area of research, providing valuable insights into some of the planet's least explored ecosystems to understand and manage marine resources. Underwater caves, in particular, hold immense potential for advancing our knowledge of archaeology, hydrology, and geology~\cite{lace2013biological}. The formations and sediments within these caves serve as critical records of historical climate and geology events while playing a key role in monitoring groundwater flow within Karst topographies, which supply freshwater to nearly a quarter of the global population~\cite{karstbook}. However, these environments are difficult to access and often hazardous for human divers due to their confined spaces, complex geometries, and lack of natural light~\cite{potts2016thirty,buzzacott2009american}. To this end, Autonomous Underwater Vehicles (AUVs) present a promising solution, offering safe and efficient exploration while minimizing risks and improving  precision~\cite{abdullah2023caveseg}.

\begin{figure}[t]
    \centering
    \includegraphics[width=\columnwidth]{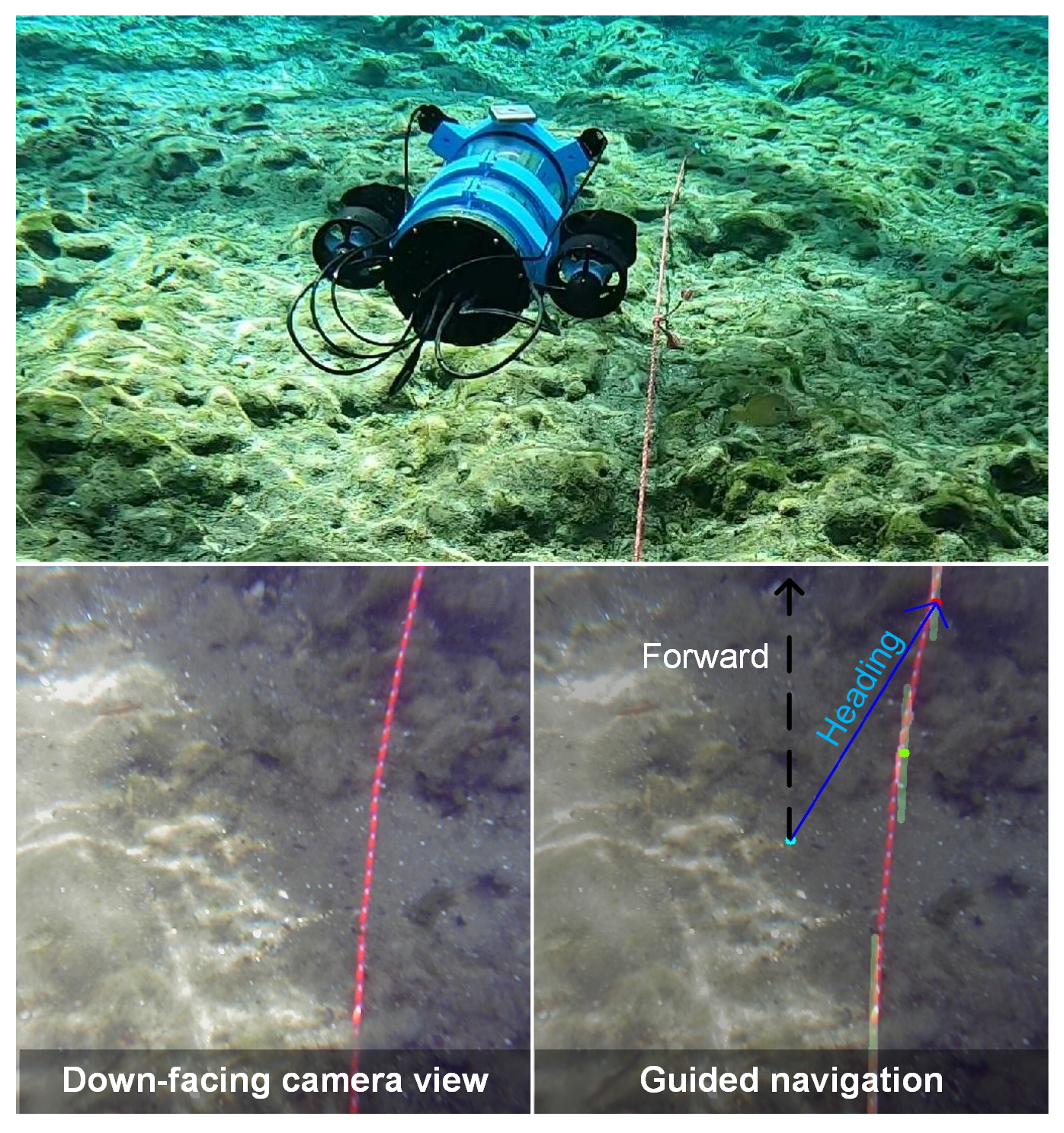}%
    \vspace{-1.5mm}
    \caption{The CavePI AUV navigates by leveraging the semantic guidance of a caveline from its down-facing camera. A deep visual perception module extracts the semantic cues, which are processed by an onboard planner to make visual servoing decisions.
    }%
    \vspace{-4mm}
    \label{fig:vehicle}
\end{figure}

However, fully autonomous exploration inside underwater caves presents significant challenges: navigation is difficult without GPS, while poor visibility, backscattering, and silt disturbances hinder perception. Thus, uninformed exploration, as often done by AUVs in open-water ecological surveying and mapping tasks~\cite{manderson2020visionbasedgoalconditionedpoliciesunderwater}, is quite challenging. For autonomous navigation inside underwater caves~\cite{richmond2020autonomous,yu2023weakly}, it is important to be aware of the entry and exit directions as well as a semantic understanding of existing navigational markers, when available. More specifically, underwater caves explored by humans are augmented by a line (string), referred to as the \emph{caveline}~\cite{exley1986basic}, which extends from the entrance through all the major sections of the cave. Other, navigational markers such as \emph{arrows} and \emph{cookies}~\cite{abdullah2023caveseg} indicate the closest path to the exit (arrows' directions) and the presence of landmarks and/or divers (cookies); real-life examples and demonstrations are provided in the the supplementary video. The caveline, together with the navigational markers, provides a 1D retraction of the 3D environment, indicating the orientation and distance from the entrance of the primary passages.

In this paper, we demonstrate the system design and algorithmic integration of CavePI, a novel AUV specifically developed for navigating underwater caves using semantic guidance from cavelines and other navigational markers; see Fig.~\ref{fig:vehicle}. 
Designed for single-person deployments, CavePI's compact design and 4-degree-of-freedom (4-DOF) motion model enables safe traversal through narrow cave passages with minimal silt disturbance. It features a forward-facing camera for visual servoing and diver-robot interaction, and a downward-facing camera for caveline detection and tracking, effectively minimizing blind spots as low as $30^{\circ}$ below the robot, that converges at a distance of $38$\,cm. A Ping sonar provides altitude data to maintain a safe distance from the ground plane. Additionally, two onboard lights enhance visibility in low-light conditions, complemented by on-device software visual filters~\cite{islam2020fast} for improved perception. The computational framework of CavePI includes two single-board computers (SBCs): a Jetson Nano for perception, and a Raspberry Pi-5 for planning and control.

For visual servoing, we integrate a caveline detector that performs real-time pixel-level segmentation of the caveline using the robot's down-facing camera stream. While prior studies have proposed vision transformer (ViT)-based frameworks such as CL-ViT~\cite{yu2023weakly} and CaveSeg~\cite{abdullah2023caveseg} for caveline detection, these approaches are computationally intensive and unsuitable for the resource-constrained SBCs. To address this, we adopt more efficient feature-extractor networks: MobileNetV3~\cite{howard2019searching} and ResNet101~\cite{he2016deep} -- combined with a lightweight DeeplabV3~\cite{chen2017rethinking} as the segmentation head. By leveraging GPU-accelerated computation on a Jetson Nano SBC, our system achieves a maximum inference rate of $18.2$ frames per second (FPS) while ensuring robust segmentation performance. The resulting segmentation map is post-processed to extract caveline contours, which are used to guide a proportional-integral-derivative (PID)-based Pure Pursuit controller~\cite{rankin1998evaluating} for precise heading adjustments for smooth tracking-by-detection~\cite{shkurti2017underwater}. 
Note that CavePI leverages the caveline position and directions for semantic guidance, while the ground plane is ranged by sonar. Other markers: arrows (exit direction) and cookies (divers' presence) trigger momentary decisions such as taking u-turns, invoking logging threads, etc. Thus, in this paper, we mainly focused on demonstrating robust caveline following performances of CavePI.

In addition, we develop a digital twin (DT) of CavePI to support pre-mission planning and testing, providing a cost-effective platform for validating mission concepts. Built using the Robot Operating System (ROS) and simulated in an underwater environment via Gazebo, it addresses the challenges posed by expensive and delicate underwater technologies that require stringent safety standards for testing. Even verified systems can exhibit unforeseen behavior due to new technological components; for example, a novel caveline detection algorithm might erroneously identify unrelated objects as cavelines, leading CavePI to follow incorrect paths and potentially creating hazardous conditions. The DT allows for prototyping, rigorous testing, and refinement of new underwater technologies, algorithms, or system modifications before physical deployment, ensuring safer and more economical evaluations. Moreover, as marine exploration missions require substantial logistical planning and time investment, initial mission planning and performance assessments conducted on the DT facilitate rapid feedback and iterative improvements, enhancing overall mission efficiency.


CavePI's guided navigation capabilities are initially evaluated by line-following experiments conducted in a controlled environment: a $2$\,m$\times3$\,m water tank with a maximum depth of $1.5$\,m. Cavelines are arranged in both regular geometric patterns (\eg, circles, rectangles) and irregular configurations such as curves, spirals, and vertical slopes. The \textit{line-following} accuracy is quantified by measuring the \textit{tracking error}, defined as the distance between the line and the optical center of the down-facing camera on the plane of the line. Similar evaluations are performed in the Gazebo simulation environment using the DT. Additionally, the delay in decision-making and its impact on an overshoot at sharp corners are analyzed. Specifically, the proportional and differential gains of the PID controller are fine-tuned via exhaustive search, and overshoots are minimized for subsequent runs. 

Following extensive in-house testing and refinement, CavePI is deployed in diverse real-world environments, including $1.5$-$6$\,m 
 deep spring waters and $12$-$30$\,m 
deep natural underwater grottos and caves. These outdoor settings pose unique perception and navigation challenges, such as strong currents and turbidity in open waters, as well as noisy, low-light conditions within overhead environments. These deployments validate the robustness of CavePI, demonstrating its capability to perform long-term autonomous missions in complex underwater environments. Our long-term goal is to make it more generalizable for exploring any overhead structures (\eg, inside ship hulls, pipelines, dams) with no predefined markers.





\vspace{1mm}
\noindent
\textbf{Need for a new AUV.} The existing AUVs, such as SUNFISH~\cite{richmond2018sunfish}, CUREE~\cite{girdhar2023curee} and UX1~\cite{martins2018ux}, offer a broad range of capabilities for marine exploration tasks. We designed CavePI specifically with (\textbf{i}) a small footprint: one person can carry and deploy it; (\textbf{ii}) a down camera dedicated for tracking navigation markers; (\textbf{iii}) a single-enclosure, low-cost, and low-power design with 4+ hours of endurance, as cave missions take longer than open-water tasks. The size, weight, cost, and power statistics of existing robots are not the ideal fit. For instance, SUNFISH (CUREE) weighs about $48$\,kg ($25$\,kg) with $151,340$\,$\mathrm{cm}^{3}$ ($63,713$\,$\mathrm{cm}^{3}$) in volume; CavePI weighs only $8.6$\,kg with $40,432$\,$\mathrm{cm}^{3}$ volume (maximum width: $38$\,cm). Furthermore, the cost of SUNFISH is hundreds of thousands of dollars; in comparison, the overall cost of CavePI is less than \$3,500. On the other hand, UX1 requires on-land logistic support and relies on a slow pendulum system for pitch control -- making it unsuitable for turbulent cave environments. BlueROV2 or other ROVs are not autonomous, and thus require significant modifications (additional cameras, autopilot, etc.) to be deployed autonomously.

%% file: src/02_Background_Related.tex
\section{Background \& Related Work}

\subsection{Autonomous Underwater Cave Exploration}
While SOTA AUVs are proficient in executing fully autonomous missions, their deployments have primarily been constrained to open water environments~\cite{sahoo2019advancements}, where obstacles are minimal. Underwater caves, however, present unique and formidable navigation challenges, including overhead and surrounding obstacles, narrow passageways, and complex topologies. Early efforts to address these challenges include the work of Mallios~\etal~\cite{mallios2016toward}, where manually operated AUVs collected acoustic data for offline mapping. Similarly, Weidner~\etal~\cite{WeidnerICRA2017,WeidnerMSc2017} utilized stereo camera systems to generate high-resolution 3D reconstructions of cave walls, floors, and ceilings.

\begin{figure*}[t]
\vspace{-4mm}
     \centering
     \includegraphics[width=0.49\linewidth]{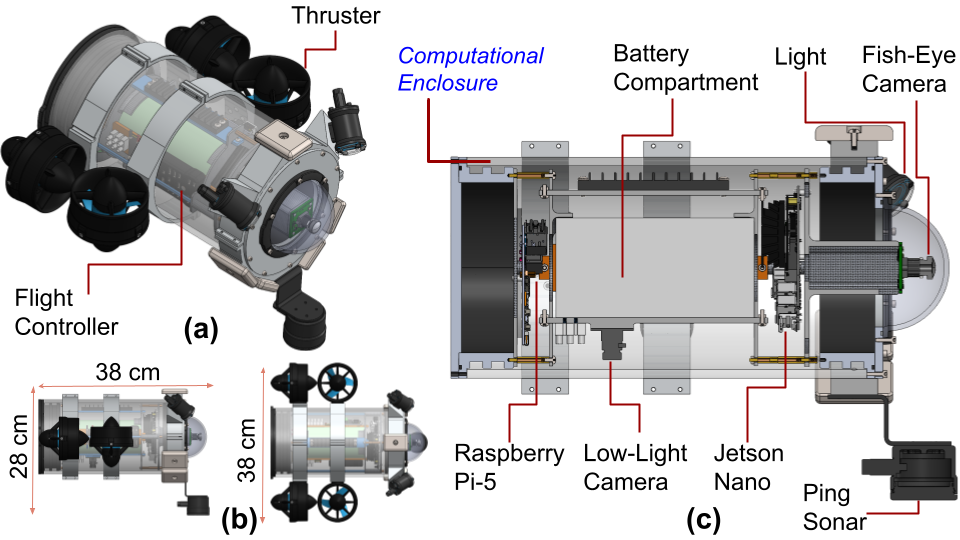}
     \includegraphics[width=0.49\linewidth]{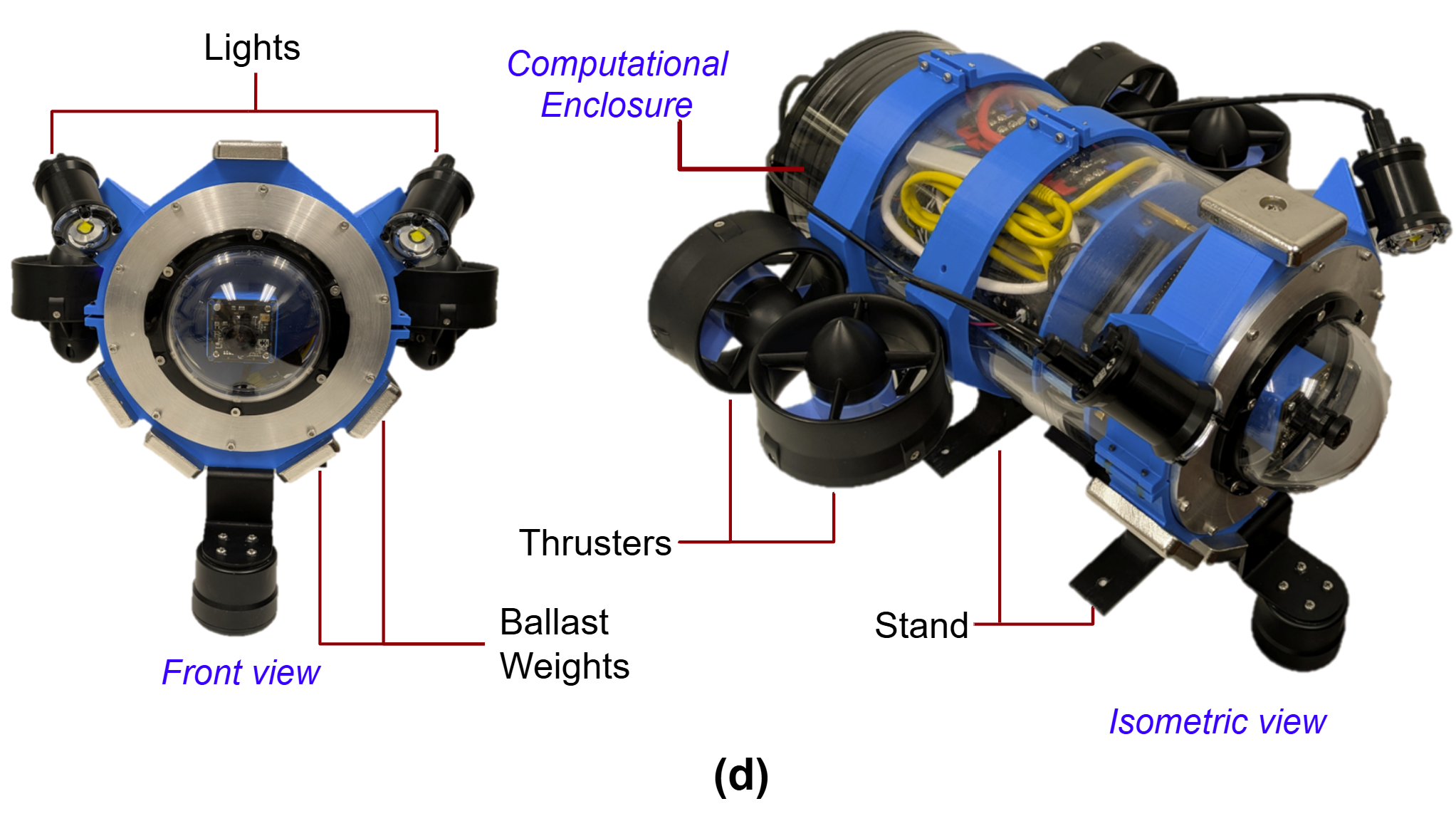}
     \caption{The proposed CavePI system design is shown; (a) isometric 3D view of the robot; (b) side-view and top-view displaying the outer shell, sonar, and thrusters' positions; (c) cross-sectional view presenting the assembly of the electronic components inside the computational enclosure; (d) the fully assembled system. CavePI is one-person deployable, weighs $8.8$\,kg, and has a depth rating of $65$ meters ($213$\,ft).}%
     \vspace{-2mm}
     \label{fig:system_design}
 \end{figure*}

Vision-based state estimation and navigation in underwater caves pose significant challenges due to factors such as lighting variability, backscattering effects, and image degradation~\cite{JoshiIROS2019}. To overcome these obstacles, Rahman~\etal~\cite{RahmanICRA2018,RahmanIJRR2022} proposed a data fusion framework that integrates visual, acoustic, inertial, and water depth measurements, enabling robust trajectory estimation and sparse representations of underwater cave environments. Advances such as shadow-based mapping~\cite{RahmanIROS2019b}, contour extraction~\cite{massone2020contour}, and real-time stereo reconstruction~\cite{WangICRA2023} have further enhanced the density and fidelity of cave mapping techniques. More recently, visual learning-based approaches have shown promise for autonomous visual servoing within underwater caves~\cite{abdullah2023caveseg,yu2023weakly}. Other contemporary works with the SUNFISH AUV~\cite{richmond2020autonomous} have demonstrated effective navigation capabilities in these environments by employing Doppler Velocity Log (DVL)-based dead reckoning and sonar-based SLAM algorithms. However, it is not one-person portable, and requires considerable logistics for successful deployment.

\subsection{Robot Navigation by Semantic Guidance}

Terrestrial and aerial robotic systems benefit from their feature-rich surroundings, leveraging semantic knowledge for navigation; examples include detecting road lanes~\cite{ding2020lane}, traffic signs~\cite{bruno2017image}, power lines~\cite{ceron2018onboard,alexiou2023visual} riverbanks~\cite{yang2022image}, and sea horizon lines~\cite{gershikov2013horizon}. Traditional approaches for extracting such features include edge detection~\cite{lee2014outdoor} and line segment analysis~\cite{yang2022image}. Advanced learning-based methods employ conditional random fields (CRFs)~\cite{zhan2020adaptive}, convolutional neural networks (CNNs)~\cite{steccanella2019deep}, and Vision Transformers (ViTs)~\cite{du2021vtnet,panda2023agronav} for robust semantic scene parsing. More recently, vision-language models, such as CLIP~\cite{radford2021learning} and its variants~\cite{shah2023lm,sontakke2024roboclip}, have been utilized to generate semantically meaningful embeddings for high-level scene comprehension and robot navigation~\cite{huang2023visual,dorbala2022clip}.

However, underwater environments are typically low-light, turbid, cluttered, and unstructured, thereby deprived of distinct semantic features~\cite{islam2024computer}. Underwater robots rely on multi-modal sensing \eg, scanning sonars for additional cues and navigation guidance~\cite{caccia2001sonar,teixeira2019dense,yu2019segmentation}. To this end, semantic knowledge representation of targets~\cite{patron2008semantic} has been proven effective for path planning and real-time decision-making of AUVs in partially-known dynamic environments~\cite{patron2010semantic}. Additionally, semantic mapping techniques using laser scanners are employed for subsea pipeline following, inspection, and intervention~\cite{vallicrosa2021semantic}. Recent studies also utilize 3D laser/sonar point clouds for underwater landmark recognition~\cite{himri2018semantic} and eventually integrate them in semantic SLAM pipelines~\cite{song2024experimental}.

For navigating underwater caves safely, learning-based frameworks have been developed for detecting and following cavelines~\cite{yu2023weakly} and other navigation markers~\cite{abdullah2023caveseg}. However, these computationally intensive models require further optimization to achieve real-time deployments~\cite{mohammadi2024edge}. Additionally, their onboard semantic segmentation performance and decision-making capabilities remain underexplored. We address this gap by conducting comprehensive real-world evaluations using our CavePI robotic platform, demonstrating its effectiveness for real-time semantic segmentation and navigation in complex underwater cave environments.



%% file: src/03_System_Design.tex
\section{CavePI System Design}
The CavePI AUV design includes three major subsystems: sensory bay, computational bay, and locomotion bay. Our proposed system and its components are shown in Fig.~\ref{fig:system_design}.  

\subsection{Sensor Bay: Acoustic-Optic Perception Subsystem}
The CavePI platform includes visual and acoustic sensors: a front-facing fisheye camera, a downward-facing low-light camera, and a Ping2 active sonar. The fisheye camera, housed within a transparent dome at the \textit{head} of the AUV, captures forward-facing visuals with a $160^\circ$ field-of-view (FOV) and outputs a video feed at $1920\times1080$ resolution.This camera detects navigation markers and visual commands (QR codes) from divers, and will assist in obstacle avoidance in the next iteration. It is worth noting that the cylindrical enclosure is a $6''$ tube while the dome has a $4''$ diameter. A custom interface is built to connect the two; see section~\ref{sub:Stability} for more details. The low-light camera, mounted inside the computational enclosure, captures downward-facing visuals with an $80^\circ \times 64^\circ$ FOV, also at the same resolution and frame rate. Additionally, a Ping2 sonar altimeter-echosounder from Blue Robotics\texttrademark{} is mounted on the underside of the robot; the sonar has a range of $100$ meters, a depth rating of $300$ meters, and a resolution of $0.5\%$ of the range, allowing it to detect obstacles directly beneath CavePI at an output frequency of $10$\,Hz. These sensory components collectively provide robust environmental awareness for autonomous navigation in challenging underwater caves.

\subsection{Computational Bay}
\vspace{-1 mm}
As illustrated in Fig.~\ref{fig:system_design}, the computational and electronic components of CavePI are housed within an acrylic cylindrical enclosure. This enclosure, with a thickness of $6.35$\,mm and a depth rating of 65 meters, forms the \textit{main body} of the robot, providing mechanical stability, buoyancy, and waterproof protection for the electronics. The computational elements include a Raspberry Pi-5, a Nvidia\texttrademark{} Jetson Nano, and a Pixhawk\texttrademark{} flight controller. The Jetson Nano is dedicated to processing visual data from the cameras, performing image processing tasks critical for scene perception and state estimation. The Raspberry Pi-5 manages planning and control modules, ensuring real-time underwater navigation. The Pixhawk flight controller acts as a bridge between hardware and software, receiving actuation commands from the Raspberry Pi-5 and transmitting them to the thrusters and lights via the {\tt MAVLink} communication protocol. Additionally, the Pixhawk integrates a 9-DOF IMU, offering 3-axis gyroscope, accelerometer, and magnetometer measurements, which are used to calculate the attitude of CavePI during underwater operations.

The enclosure also contains the battery compartment, voltage regulators, electronic speed controllers (ESCs), and a Bar-30 pressure sensor. The battery compartment holds a $14.8$\,V ($18$\,Ah) battery pack, regulated to power internal components (\eg, cameras, computers) and external components (\eg, thrusters, sonar). The battery pack is capable of supplying sufficient power to sustain over 6 hours of operation of CavePI. Each thruster is controlled by an ESC, which drives the three-phase brushless motor using PWM signals from the Pixhawk. The Bar-30 sensor provides high-precision pressure readings with a resolution of $0.2$\,mbar and an accuracy of $2$\,mm, with a working depth of up to $300$ meters. This pressure data is processed to determine underwater depth, ensuring reliable and accurate interoceptive perception during operations.

\begin{figure}[t]
    \centering
    \includegraphics[width=0.98\linewidth]{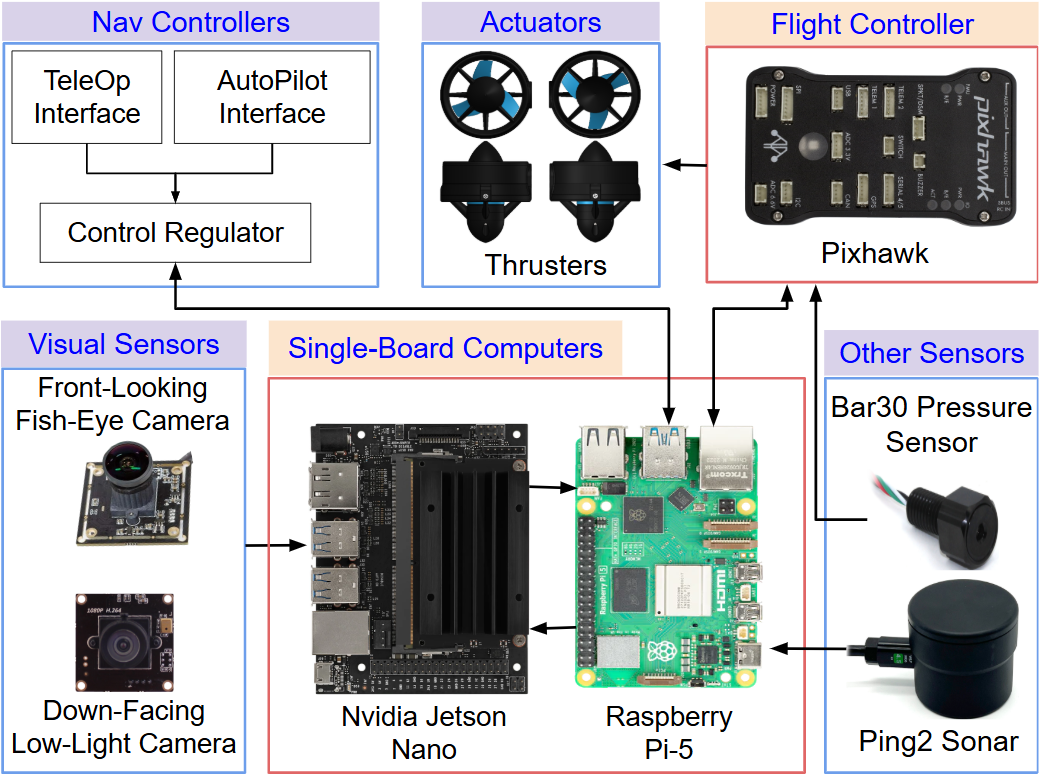}%
    \vspace{-1mm}
    \caption{Major electronics and sensor-actuator connections of CavePI.}
    \label{fig:electronics}
    \vspace{-3mm}
\end{figure}

\subsection{Locomotion Bay: Middleware Integration}
\vspace{-1 mm}
The end-to-end integration of CavePI ensures that each computational component operates in sync, tied to a ROS2 Humble-based middleware backbone. The modular design also allows for future upgrades, ensuring that the CavePI can be tailored to meet evolving research in marine ecosystem exploration and monitoring. The sensor-actuator signal communication graph is illustrated in Fig.~\ref{fig:electronics}.

\begin{figure}[t]
    \centering
    \includegraphics[width=\linewidth]{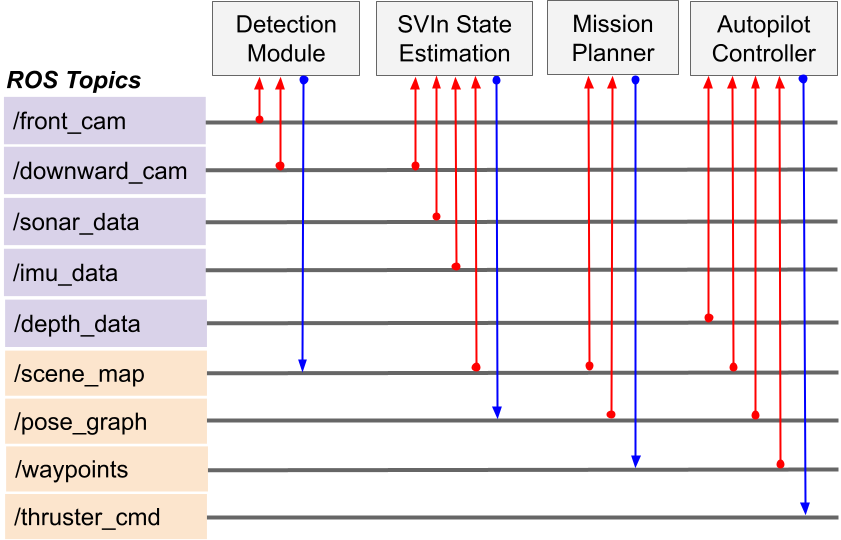}%
    \vspace{-1mm}
    \caption{Data flow among major computational modules of CavePI is shown in the form of \texttt{ROS Topics}: red and blue arrows represent \textit{subscribed} and \textit{published} topics in the ROS graph, respectively.}
    \label{fig:ROS}
    \vspace{-4mm}
\end{figure}

The CavePI AUV is designed for low-power operation and integrates a modular ROS2 framework to support application-specific perception, planning, and control methods. As depicted in Fig.~\ref{fig:ROS}, the \textit{detector} node acquires visual data from the two cameras to identify the caveline for navigation. The \textit{mission planner} node then integrates the caveline information with the estimated position data to generate subsequent waypoints for the mission. Finally, the \textit{autopilot controller} node utilizes these waypoints, along with the detected caveline, positional data, and depth readings from the Bar-30 sensor, to generate precise actuation signals for the thrusters, enabling accurate movements and depth control. Additionally, CavePI can function as an ROV through an optional tether-based teleoperation module. This module transmits user commands from a joystick to the onboard Raspberry Pi-5, which processes the inputs and relays them to the thrusters for manual control.

\vspace{1mm}
\noindent
\textbf{Power footprint and synchronization.} As shown in Table~\ref{tab:battery_power_consumption_table}, CavePI offers an endurance of 6+ hours at maximum capacity. While it can operate much longer, the battery profile~\cite{BR_battery} recommends running it above $20$\% power; the remaining capacity tends to deplete rapidly, which is a safety issue for recovery. While running in full capacity, the integrated system ensures an overall system throughput of $3.6$\,Hz. Specifically, both the cameras run at $30$\,Hz; the IMU and sonar frequencies are $100$\,Hz and $10$\,Hz, respectively. In our implementation, software synchronization is achieved via a joint ROS2 publisher that shares a clock across all sensory topics. 

\begin{table}[h!]
\centering
\caption{
The battery power consumption characteristics of CavePI (at maximum capacity) over time, showing 6+ hours of endurance.
}
\footnotesize
\begin{tabular}{c|ccccccc}
\toprule
\textbf{Time (hours)} & 1 & 2 & 3 & 4 & 5 & 6 & 7 \\
\midrule
\textbf{Battery \%} & 85 & 73 & 61 & 50 & 39 & 29 & 19 \\
\bottomrule
\end{tabular}
\label{tab:battery_power_consumption_table}
\end{table}

\subsection{CavePI Digital Twin}
We develop a digital twin (DT) model of CavePI by using the Unified Robot Description Format (URDF), with links and joints carefully assigned to represent the various CAD components designed in SolidWorks. To replicate the sensor suite of the physical CavePI, Gazebo plugins are integrated to simulate the front-facing camera, down-facing camera, IMU, pressure sensor, and sonar. Additional plugins are employed to simulate environmental forces, including buoyancy, thrust, and hydrodynamic drag, thereby enhancing the physical realism.

A controlled open-water scenario is created in Gazebo to simulate realistic missions, featuring a thin line arranged in a rectangular loop to mimic a caveline. Since the simulated environment lacks real-world perception challenges such as low light or turbid water conditions, the perception subsystem remains simplified. Instead of deploying computationally intensive deep visual learning models, simpler edge detection and contour extraction techniques~\cite{SUZUKI198532} are used to identify the caveline from the down-facing camera feed. The remaining navigation and control subsystems mirror the real-world implementation and operate via two ROS nodes. The first node processes the extracted contours to make navigation decisions and publishes high-level control commands (\eg, yaw angle). The second node subscribes to these commands, computes the required thrust and hydrodynamic drag forces, and publishes them as ROS topics to control the simulated robot model.

Beyond replicating caveline following experiments, we utilize the DT system for preliminary testing and fine-tuning of new control algorithms. It also enables the simulation of complex cave environments, such as narrow passages, dead ends, and sharp turns. Conducting repeated real-world experiments in such scenarios to improve the control system can be logistically demanding where the simulation offers an efficient alternative for extensive evaluation and fine-tuning.  




%% file: src/04_Nav_Pipeline.tex
\section{CavePI Navigation Pipeline}
The CavePI AUV is designed to autonomously navigate underwater by following a caveline and other navigation markers. However, a caveline appears a few pixels wide in the bottom-facing camera’s FOV and is significantly challenging to detect in noisy conditions with almost no ambient light. Moreover, the caveline is often obscured or blends into the background, requiring a robust visual learning pipeline for reliable operation. 

\subsection{Semantic Guidance: Tracking by Detection}
The model selection process plays a crucial role in optimizing accuracy and efficiency on resource-constrained edge devices. While object detection and semantic segmentation are two established techniques in visual learning, they produce different interpretations of the target RoI. Object detection models generate bounding boxes around the target, which, given the caveline's irregular shape and orientation, can encompass significant background regions, complicating the estimation of the heading angle. To address this, we opt for semantic segmentation, which provides pixel-level contours of the caveline as a more precise marker for tracking-by-detection.

\begin{figure}[h]
    \centering
    \includegraphics[width=\linewidth]{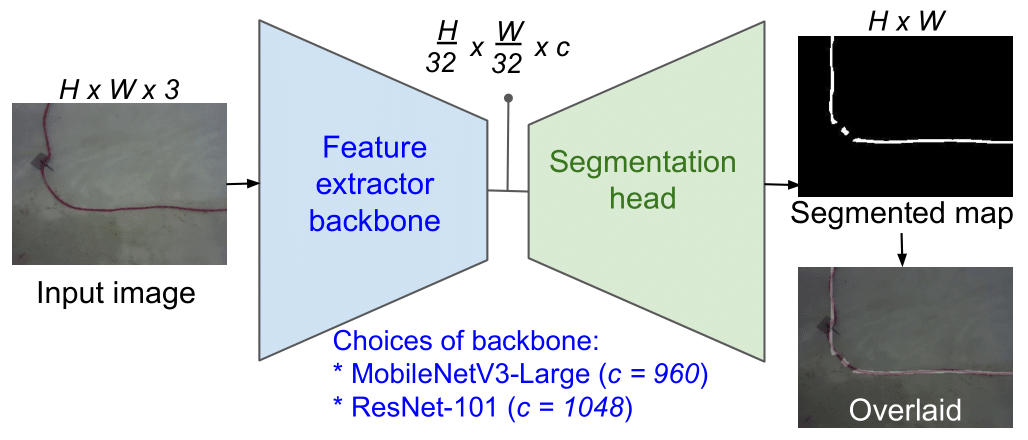}%
    \caption{Simplified model architecture for caveline segmentation is shown; we use a DeepLabV3~\cite{chen2017deeplab} head with two choices for backbone network: MobileNetV3~\cite{howard2019searching} and ResNet101~\cite{he2016deep}.}
    \label{fig:model}
\end{figure}

\vspace{1mm}
\noindent
\textbf{Model architecture and design choices.} Onboard resource constraints of CavePI significantly influence the choice of model architecture. The Jetson Nano, with its shared memory architecture, allocates GPU memory dynamically from a limited $2$\,GB pool, which must also accommodate the dual camera feeds and Nano-Pi control signal communication. Considering these limitations, we initially selected the MobileNetV3 (Large) backbone~\cite{howard2019searching} for feature extraction. This lightweight architecture feeds extracted features into a DeepLabV3~\cite{chen2017rethinking} head to generate binary segmentation maps. While the MobileNetV3-DeepLabV3 combination is computationally efficient, requiring only about $314$\,MB GPU memory for $11$ million parameters, it proved insufficient for accurately segmenting the thin caveline in turbid water conditions. To address this limitation, we adopted a heavier backbone, ResNet101~\cite{he2016deep}, while retaining the DeepLabV3 head. This model achieved slightly better performance, as highlighted in Table~\ref{tab:model_comparison}, albeit at the cost of reduced inference speed. Both architectures were initialized with pre-trained weights and fine-tuned on a custom dataset to enhance caveline detection accuracy in challenging environments.

\vspace{1mm}
\noindent
\textbf{Dataset and model training.} 
For this study, we utilized the open-source CL-ViT dataset, which includes pixel-level annotations of cavelines~\cite{yu2023weakly}. It comprises $3,150$ RGB images collected from three distinct underwater cave systems. To enhance the dataset's diversity and variance, we manually annotated an additional $150$ images from our laboratory testbed, resulting in a combined dataset of $3,300$ images. The dataset was divided into $3,200$ training and $100$ validation images. Both models were fine-tuned on this dataset over a maximum of $50$ epochs using the following training configurations: the Adam optimizer with an initial learning rate of $0.001$, a cosine learning rate scheduler with $T_{max} = 20$, and a cross-entropy loss function. This training configuration was designed to optimize these models' performances in diverse underwater environments.

\vspace{1mm}
\noindent
\textbf{Benchmark comparison.} The fine-tuned models are evaluated on the open-source CL-Challenge dataset~\cite{yu2023weakly} that contains caveline annotations for $200$ noisy low-light images from different cave systems. Using the lighter MobileNetV3 backbone, our model achieves a mean intersection-over-union (mIoU) of $48.95\%$, whereas the considerably heavier ResNet101 model demonstrates only a marginal improvement with a mIoU of $50.48\%$. For reference, the highest mIoU score reported on this dataset is $58.3\%$~\cite{yu2023weakly}. Since smooth real-time operation requires a minimum control loop of $3.6$\,Hz, we opt for the MobileNetV3 backbone that exceeds this requirement with an inference speed of $18.2$\,fps (frames per second).

\begin{table}[h]
\vspace{1mm}
    \centering
    \renewcommand{\arraystretch}{1.2}
    \caption{Edge performances (on a Jetson Nano device) for the two model configurations (Fig.~\ref{fig:model}) available in CavePI are compared.}
    \resizebox{\linewidth}{!}{
    \begin{tabular}{lcccc}
    \Xhline{2\arrayrulewidth}
    Backbone & mIoU & \# Params & Inference rate & GPU usage\\ 
    \Xhline{2\arrayrulewidth}
    MobileNetV3~\cite{howard2019searching}  & $48.95\%$ & $11.02$\,M & $18.2$ fps  & $314$\,MB  \\
    ResNet101~\cite{he2016deep}             & $50.48\%$ & $60.99$\,M & $\z1.4$ fps & $747$\,MB  \\
    \Xhline{2\arrayrulewidth}
    \end{tabular}
    }
    \label{tab:model_comparison}
    \vspace{-1mm}
\end{table}

\vspace{1mm}
\noindent
\textbf{Model compression and integration.} The two aforementioned models are optimized to deploy on the Jetson Nano device of CavePI. First, the models are converted into Open Neural Network Exchange (ONNX)~\cite{onnx} format that uses computational graphs (nodes and edges) to represent operations and data flow within the pipeline. Subsequently, a serialized data structure is created by NVIDIA's TensorRT SDK~\cite{tensorrt}, which stores highly optimized deep learning models ready for fast inference on NVIDIA GPUs. Table~\ref{tab:model_comparison} compares the onboard performance of the two models used in this work.

\subsection{Autonomous Control} 
\begin{figure}[t]
    \centering
    \includegraphics[width=\linewidth]{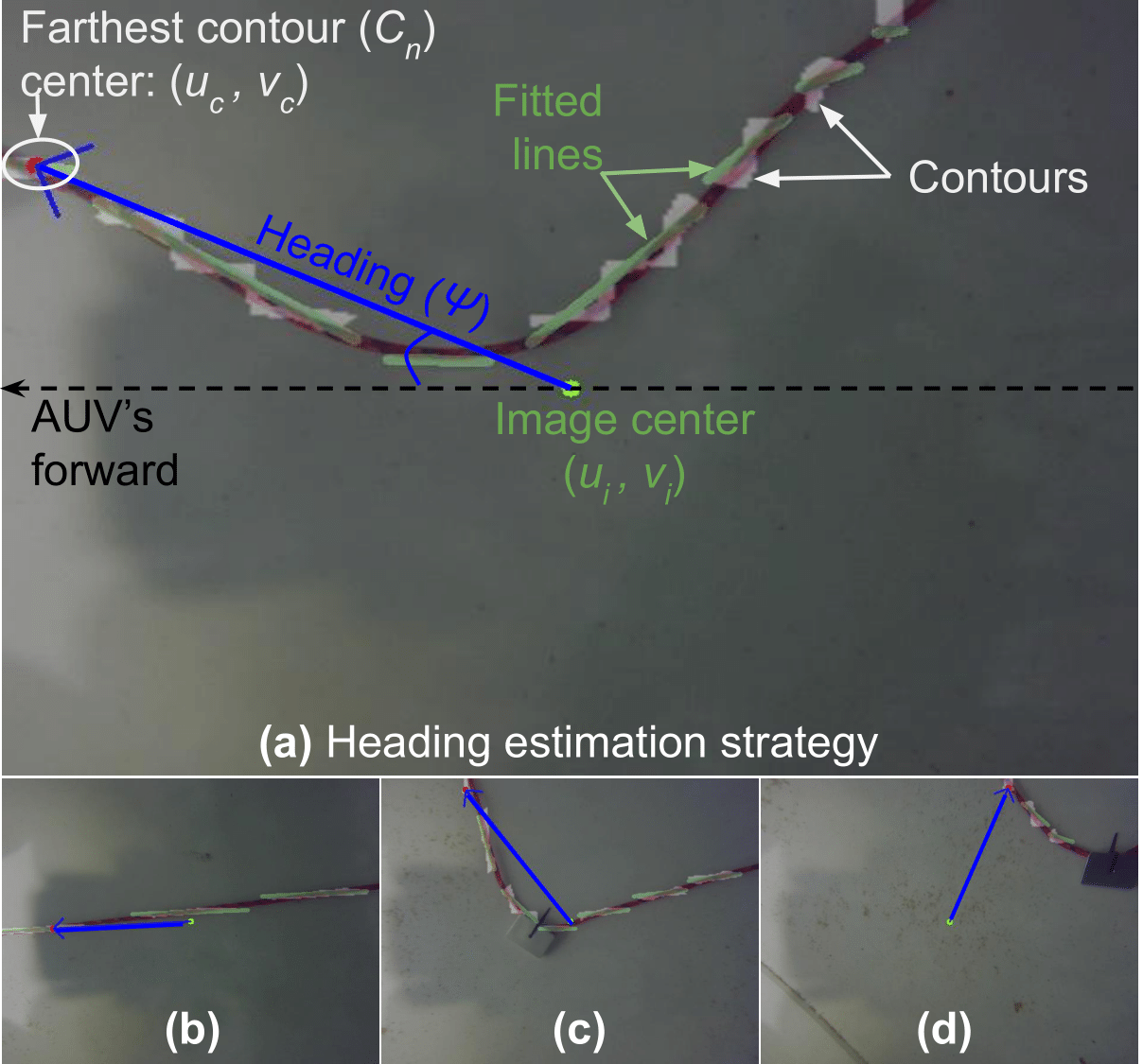}%
    \vspace{-1mm}
    \caption{A few visual servoing test cases are shown; (a) AUV heading is estimated based on its farthest caveline contour; (b,c) scenarios for straight and turn decisions, respectively; and (d) an \textit{overshoot} scenario occurred due to delayed decision-making.
    }
    \label{fig:heading_control}
    \vspace{-4mm}
\end{figure}

Algorithm~\ref{alg:line_follower} outlines the visual servoing control algorithm employed by CavePI. The process begins by extracting caveline contours, $\mathcal{C}$, from the segmentation map $\mathcal{I}$, for semantic guidance. As the caveline is often detected as fragmented contours (see Fig.~\ref{fig:heading_control}), the center of the farthest contour $(u_c, v_c)$ is identified and selected as the next waypoint in CavePI's path planning. The heading angle $\psi$ of this waypoint is then calculated with respect to the image center $(u_i, v_i)$, about the x-axis of the image frame, following the right-hand rule. This heading angle serves as a high-level navigation command, which is transmitted to the Raspberry Pi-5 for execution within the computational subsystem.

\begin{algorithm}[t]
\caption{Visual Servoing Controller of CavePI}
\label{alg:line_follower}
\begin{algorithmic}[1]
\STATE \textbf{Input:} Segmentation map $\mathcal{I}_{w\times h}$, center pixel $(u_i, v_i)$
\STATE \textbf{Output:} Thruster commands $\mathbf{v}$

\WHILE{robot is active}
    \STATE Extract set of contours $\{\mathcal{C}\}$ from $\mathcal{I}$
    \IF{$\mathcal{C} = \emptyset$}  
        \STATE \textbf{State} $\gets$ \texttt{lost} \codecomment{no caveline detected}
        \STATE Rotate $360^\circ$ \codecomment{small loops for recovery} 
    \ELSE
        \IF{$|\mathcal{C}| > 1$} 
            \STATE Sort $\{\mathcal{C}\}$:~$\{\mathcal{C}_0, \mathcal{C}_1, ..., \mathcal{C}_n, ...\}$ \codecomment{toward robot's heading}
            \STATE $(u_n, v_n) \gets \text{Centroid} (\mathcal{C}_n)$
            \STATE $\mathcal{C}_\text{next} \gets \arg\max_{\mathcal{C}_n \in \{\mathcal{C}\}} \bigl((u_n - u_i)^2 + (v_n - v_i)^2\bigr)$ \codecomment{farthest contour from $(u_i,v_i)$}

        \ELSE
            \STATE $\mathcal{C}_\text{next} \gets \mathcal{C}$
        \ENDIF
        \STATE $(u_c, v_c) \gets \text{Centroid}(\mathcal{C}_\text{next})$ \codecomment{contour center}
        \STATE $\psi \gets \text{slope}((u_i, v_i), (u_c, v_c))$ \codecomment{next heading angle}
        \STATE $\mathbf{v} \gets f(\psi)$ \codecomment{thruster commands}
    \ENDIF
    \STATE Send $\mathbf{v}$ to thrusters
\ENDWHILE
\end{algorithmic}
\end{algorithm}

The centroid of all detected contours in the set $\mathcal{C}$ is calculated, with the centroid of the $n^{\text{th}}$ contour denoted as \((u_n, v_n)\). The farthest contour is determined by:
\begin{equation}
(u_c, v_c) = \arg\max_{n \in C}\, \big[{(u_n - u_i)^2 + (v_n - v_i)^2}\big].
\end{equation}
The instantaneous heading vector is defined as the vector from the image center \( (u_i, v_i) \) to the centroid of the farthest detected contour \( (u_c, v_c) \), represented as $\begin{bmatrix} u_c - u_i & v_c - v_i \end{bmatrix}^T$. The equation of the instantaneous line of motion in the image frame is given by:
\begin{equation}
\begin{bmatrix} u - u_i \\ v - v_i \end{bmatrix} \times \begin{bmatrix} u_c - u_i \\ v_c - v_i \end{bmatrix} = 0,
\end{equation}
where \( (u, v) \) represents the free variable on the line.

Ideally, the forward x-axis of the image frame aligns with the robot's heading direction, enabling the computed heading angle to function as the primary error signal for a PID controller on the Raspberry Pi-5. The PID controller is calibrated to minimize this error, maintaining CavePI’s heading toward the designated waypoint. The controller-generated signals are transmitted via the {\tt MAVLink} communication protocol to the electronic speed controllers (ESCs), which regulate the speed and rotational direction of the thrusters.
Upon failure to detect any caveline, CavePI executes a circular search pattern to re-acquire tracking. If a line is not detected, it signals flashing lights (helps if divers are nearby); subsequently, it starts navigating towards the (last seen) arrow direction (toward the cave exit) to rediscover the caveline and/or exit the cave. Additionally, CavePI maintains a user-defined target depth by utilizing a PID controller, which processes the error betweein the current and desired depths and generates appropriate control signals to achieve stable depth regulation.

%% file: src/05_System_Evaluation.tex
\section{System Evaluation}

\subsection{Mechanical Stability in Hydrostatic Pressure}
\label{sub:Stability}
The CavePI system is evaluated for structural robustness and functional durability to support long-term autonomous operation at depths. Critical components of AUVs generally include the interfacing joints of the actuators (thrusters), sensors (\eg, cameras and sonar), and enclosure doors~\cite{macias2024numerical}. As shown earlier in Fig.~\ref{fig:system_design}, CavePI’s \textit{head} section is connected to the \textit{main body} via a custom-designed aluminum plate, referred to as the dome connector, which ensures a watertight interface for the forward-facing camera. Similarly, the clamps used in the thruster and sonar subassemblies, as well as the plates for electronics mounting within the computational enclosure, are custom-designed to enhance mechanical efficiency and dynamic stability. Components exposed directly to the underwater environment, including thrusters, the Ping2 sonar, the dome connector, and external cables, are identified as sensitive elements requiring careful design and validation to ensure system reliability under challenging conditions.

\begin{figure}[t]
     \centering
     \includegraphics[width=0.96\columnwidth]{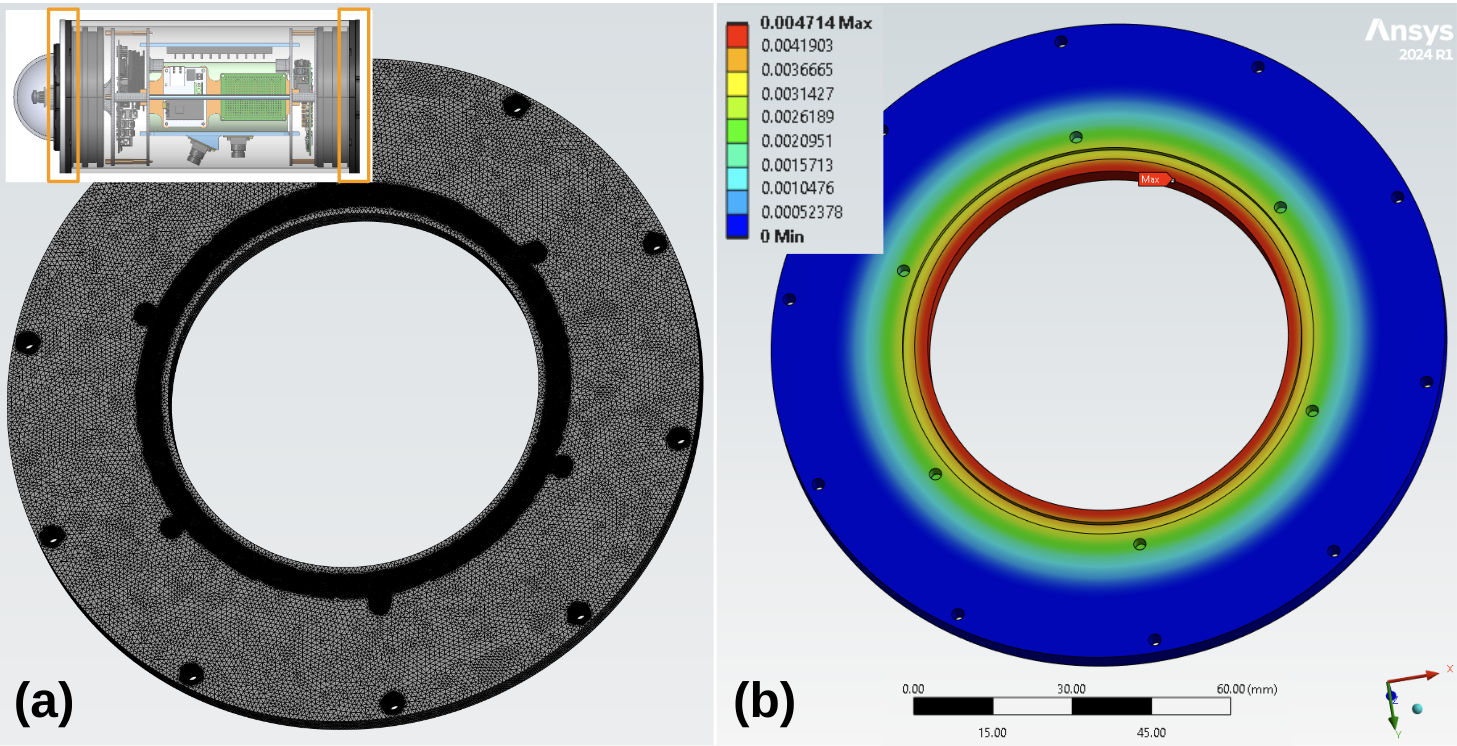}
     \includegraphics[width=0.96\columnwidth]{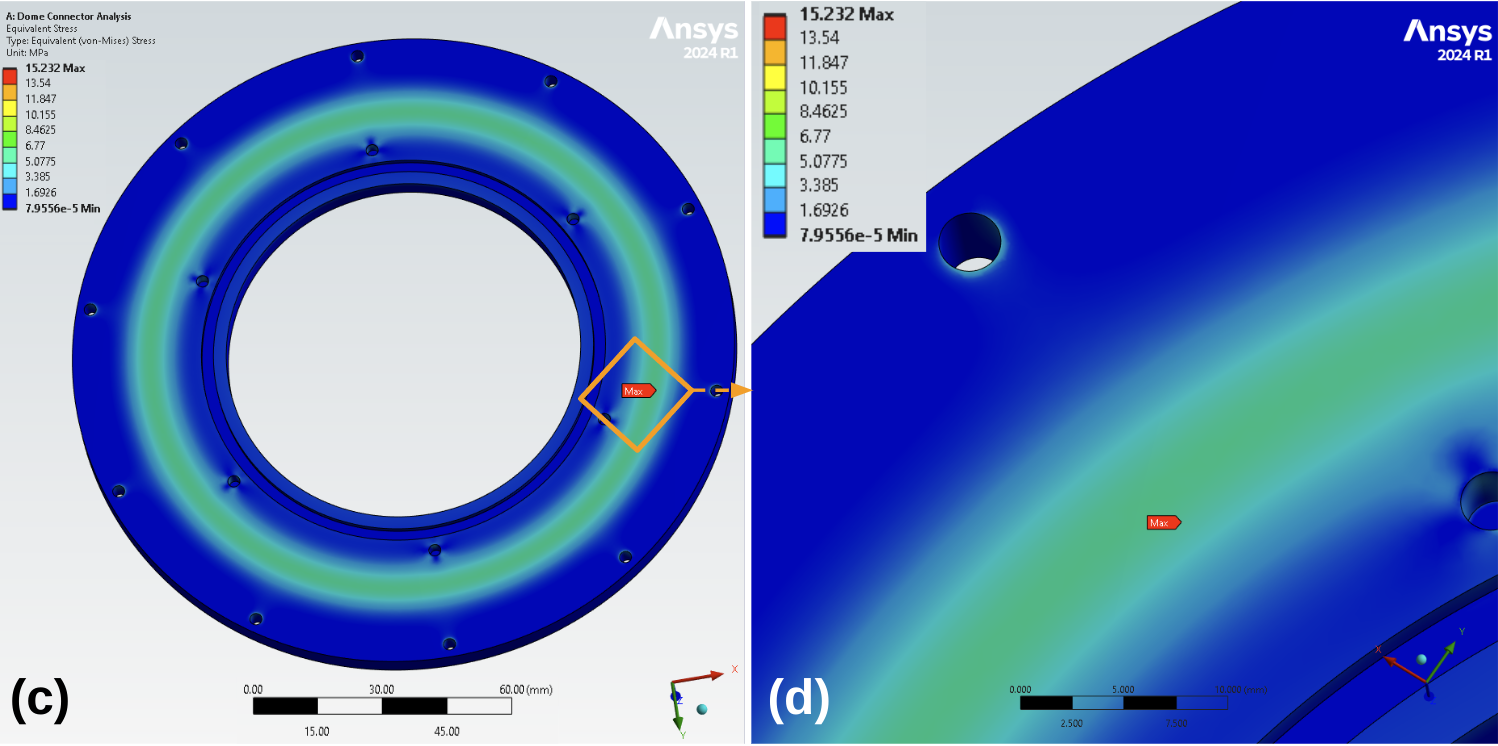}%
     \vspace{-1mm}
     \caption{(a) FEA mesh of the dome connector; (b) its total deformation due to hydrostatic forces at $65$\,m water depth; (c) equivalent stresses as per Von-Mises yield criterion; and (d) maximum stresses at the zoomed-in area (best viewed digitally at $2\times$ zoom).}%
     \vspace{-2mm}
     \label{fig:dome_connector_stress}
 \end{figure}

\begin{figure*}[t]
     \centering
     \includegraphics[width=\linewidth]{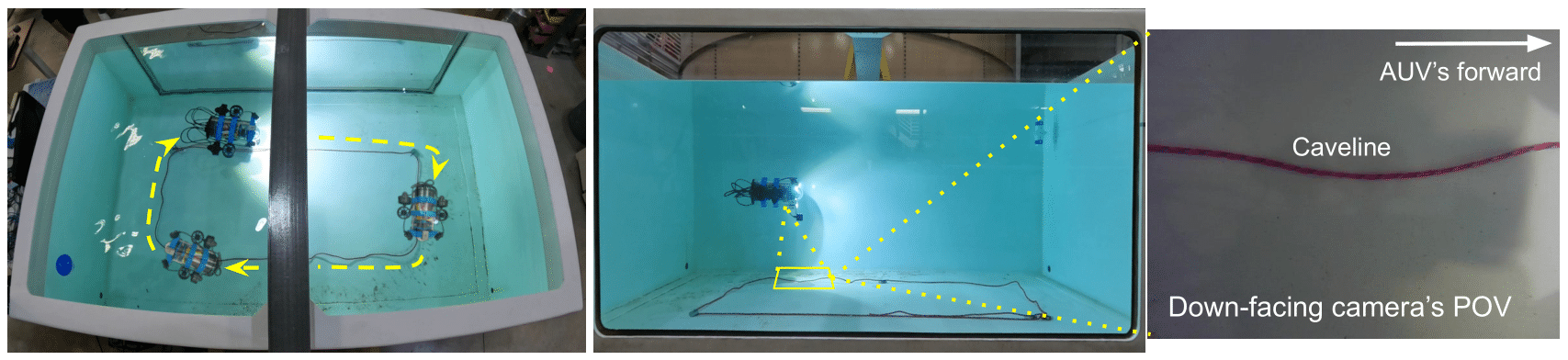}%
     \vspace{-1.5mm}
     \caption{The laboratory setup used for tracking accuracy evaluation is shown. CavePI detects and follows the line laid on the tank floor, completing a loop around the $6$\,meter perimeter in approximately $2$\,minutes.}%
     \vspace{-4mm}
     \label{fig:tank_setup}
 \end{figure*}

The thrusters and Ping2 sonar have depth ratings of more than $500$ meters and $300$ meters, respectively, along with their cables. To evaluate the structural strength of the dome connector of CavePI, we use finite element analysis (FEA) methods~\cite{bieze2018finite}. Tetrahedral elements are utilized to generate the finite element mesh from the 3D model of the dome connector, while the Von Mises yield criterion is applied to assess the \textbf{maximum stresses}, \textbf{total deformation}, and \textbf{factor of safety ({\tt FoS})}. The analysis considered hydrostatic pressures~\cite{macias2024numerical,li2023soft} at the maximum water depth of $65$\,m.

The number of nodes and finite elements in the mesh indicates the mathematical model's quality: a higher count typically reflects a more accurate model. In the analysis of the dome connector, the mesh contained $4,636,236$ nodes and $3,300,641$ elements; see Fig.~\ref{fig:dome_connector_stress}. Based on the Von-Mises criterion, the maximum equivalent stress was calculated to be $15.3$\,MPa. The maximum deformation, observed at the inner most circumference of the dome connector, was $0.005$\,mm. The minimum {\tt FoS} is calculated using the yield strength ($\sigma_y$) of the material (Aluminium 6061-T6), and maximum stress ($\sigma_{max}$) as follows: 
\begin{equation*}
    \text{{\tt FoS}} = \frac{\sigma_y}{\sigma_{\text{max}}} = \frac{240 \text{ MPa}}{15.3 \text{ MPa}} \approx 15.76.
\end{equation*}
Note that a {\tt FoS} greater than $2$ for hydrostatic applications and $4$ for hydrodynamics applications is considered sufficient~\cite{kazemi2004reliability,SafetyCulture}. Thus, an {\tt FoS} over $15$ indicates that the dome connector is sufficiently stable for the intended operation pressure.

\subsection{Caveline Following Performance}\label{sec_5b}
\noindent
\textbf{Setup.} The line following experiments are initially conducted in a  $2$\,m$\times3$\,m laboratory water tank with a maximum depth of $1.5$\,m. 
As a first setup, a line is laid on the tank floor in a rectangular loop, without any depth variation; see Fig.~\ref{fig:tank_setup}. Subsequently, we add and vertical slopes in the tank to simulate the rough underwater terrains; see Fig.~\ref{fig:all_setup}\,(a). Upon detecting the line via the down-facing camera, CavePI autonomously follows the loop while maintaining a specified depth. The system's line-following accuracy, depth-holding capabilities, and control smoothness are evaluated and fine-tuned in preparation for outdoor deployments.

\begin{figure}[ht]
\centering
\begin{subfigure}[]{0.9\linewidth}
\includegraphics[width=\linewidth]{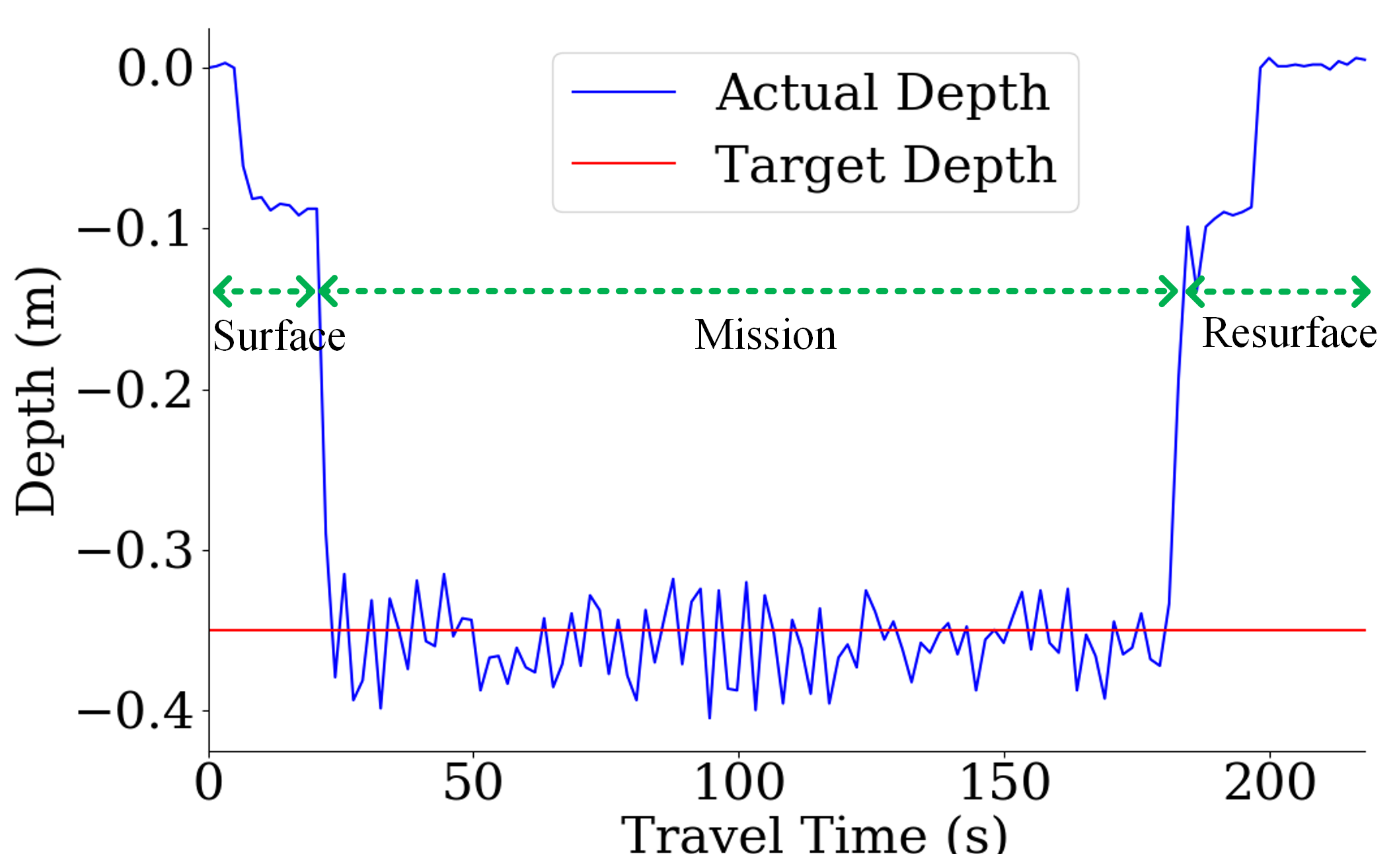}%
\caption{Depth Control Accuracy}
\end{subfigure}
\begin{subfigure}[]{0.9\linewidth}
\includegraphics[width=\linewidth]
{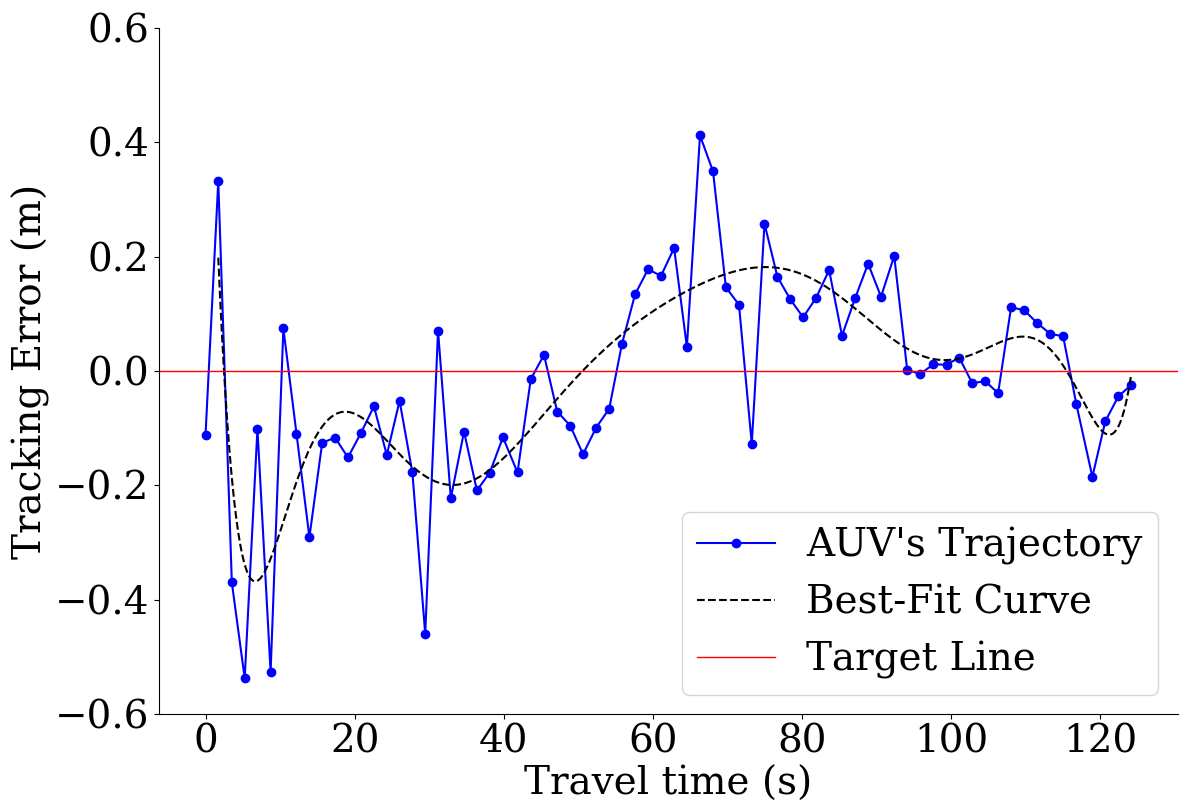}%
\caption{Tracking Accuracy}
\end{subfigure}
\caption{Line-following and depth-holding accuracy are reported for a laboratory experiment: (a) The target depth was set to $0.35$\,m; (b) the tracking error ($\delta$), \ie, the offset of the optical center of the camera from the detected line is plotted for a rectangular loop following task.  
}
\label{fig:deviation}
\vspace{-4mm}
\end{figure}

 \begin{figure}[t]
    \centering
    \includegraphics[width=\columnwidth]{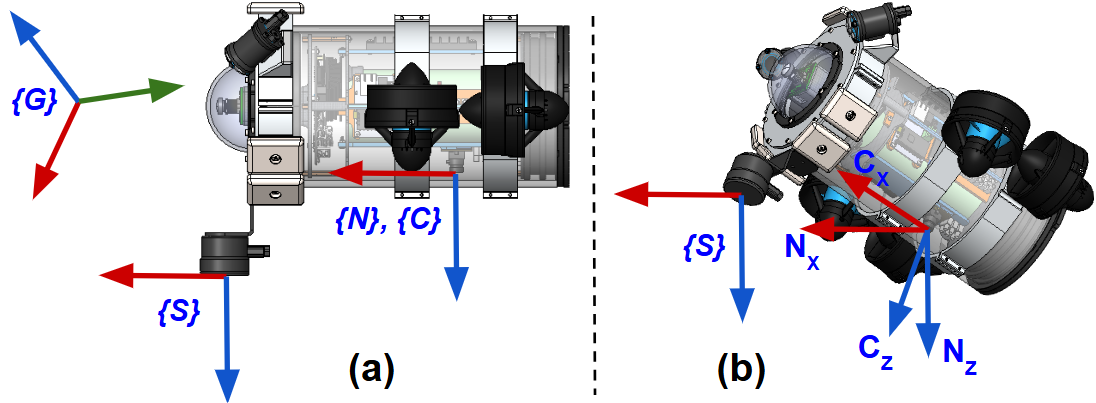}%
    \vspace{-1mm}
    \caption{Configurations of the coordinate frames are shown at: (a) initial time $t_0$, and (b) an arbitrary time $t$. Note that the \{C\} is a (rigid) body-fixed frame attached to the robot while \{N\} and \{S\} are non-rotating (fixed orientation) frames, even though their origins translate along with the robot.
    }%
    \vspace{-2mm}
    \label{fig:frames}
\end{figure}

\vspace{1mm}
\noindent
\textbf{Evaluation process.}  CavePI's caveline-following performance is evaluated by measuring the \textit{tracking error}, defined as the perpendicular distance between the optical center of the downward-facing camera and the nearest point on the caveline in the plane of caveline; see Fig.~\ref{fig:deviation}. This measurement leverages pose information from CavePI's IMU, depth readings from the sonar, and visual information from the segmentation mask of the caveline captured by the down-facing camera.

To formalize the calculation, we define several reference frames; see Fig.~\ref{fig:frames}. The initial pose of the robot assumes its front axis is aligned with the caveline and its downward axis perpendicular to the caveline plane. This reference frame, denoted as \{N\}, has its origin at the camera. A corresponding frame \{S\}, with the same orientation as \{N\} but with its origin at the sonar, is also defined. Additionally, we define the global reference frame \{G\}, and the CavePI body frame \{C\} with its origin at the down-facing camera. The transformation matrix from any frame $a$ to $b$ is defined as ${}^{b}_{a}\mathbf{T} = [ {}^{b}_{a}\mathbf{R}_{3\times3} \,|\, {}^{b}_{a}\mathbf{t}_{3\times1}]$. The roll, pitch, and yaw angle measurements from IMU are expressed in the global frame \{G\} at any time instance.

For a given image $I$, we extract the caveline contour closest to the camera center and calculate the distance between the camera center and the pixel ${}^IP$ on the contour edge, then convert it to a 3D point ${}^C\boldsymbol{P}$ using:

\begin{equation}
  {}^I{P} = 
  \begin{bmatrix}
      u_{P} & v_{P} & 1
  \end{bmatrix}^T, 
  \quad
  {}^C\boldsymbol{P} = \lambda \, K^{-1} ({}^IP);
\end{equation}
where $K$ is the camera intrinsic matrix and  $\lambda$ is the depth scale factor.
The rotation of the CavePI's frame with respect to the sonar frame is calculated as follows:
\begin{equation}
      {}^{C}_{S}\boldsymbol{R} 
      = \left({}^{S}_{G}\boldsymbol{R}\right)^{-1} {}^{C}_{G}\boldsymbol{R} 
      = \left({}^{N}_{G}\boldsymbol{R}\right)^{-1} {}^{C}_{G}\boldsymbol{R};
    \end{equation}
where ${}^{C}_{G}\boldsymbol{R}$ and ${}^{N}_{G}\boldsymbol{R}$ are computed from instantaneous and initial IMU measurements, respectively. 
Subsequently, ${}^C\boldsymbol{P}$ is transformed to the sonar frame using:   
\begin{equation}
      \begin{bmatrix}
          {}^S\boldsymbol{P} \\
          1
      \end{bmatrix}
      =
      \begin{bmatrix}
          {}^{C}_{S}\boldsymbol{R} & {}^{C}_{S}\boldsymbol{t} \\
          0 & 1
      \end{bmatrix}
      \begin{bmatrix}
          {}^C\boldsymbol{P} \\
          1
      \end{bmatrix},
    \end{equation}
    
    \begin{equation}
      {}^S\boldsymbol{P} 
      = \lambda
      \begin{bmatrix}
          a & b & c
      \end{bmatrix}^T;
    \end{equation}
where the scalars \(a\), \(b\), and \(c\) are obtained from the above equation. The raw depth measurement from sonar ${}^Cd$ is converted to the sonar frame ${}^Sd$ and $\lambda$ is calculated as:
    \begin{equation}    
        {{}^{S}d} = {}^{C}_{S}\boldsymbol{R}_{(3,3)} \cdot {}^{C}d, ~ \text{and}~\lambda = \frac{{}^Sd}{c}.
    \end{equation}
    
The 3D coordinates of ${}^N\boldsymbol{P}$ is calculated using:
    \begin{equation}
      {}^N\boldsymbol{P} = -{}^{C}_{S}\boldsymbol{t} + {}^S\boldsymbol{P}.
    \end{equation}
Finally, $\delta$ is obtained from the \(y\)-coordinate of \({}^N\boldsymbol{P}\).

\vspace{1mm}
\subsection{PID Controller Tuning}
To ensure stable operation in turbulent water conditions, CavePI’s PID-based \textit{Pure Pursuit} controller~\cite{rankin1998evaluating,scharf1969comparison} is designed and fine-tuned through extensive experimentation. Pure Pursuit is a widely used geometric tracking algorithm that guides a vehicle toward a target point on a path~\cite{ samuel2016review} at a fixed \textit{lookahead distance}. By steering the vehicle to continuously align with this moving target point, the controller ensures smooth convergence to and following of the intended trajectory, which we tuned and tested empirically.

Our initial trials reveal challenges, including high overshoot, difficulty maintaining depth, and instability at sharp corners. These issues are iteratively calibrated to achieve robust and reliable motion control. A grid search technique is employed to optimize the proportional gain ($K_p$) and differential gain ($K_d$) for the two onboard controllers -- depth and heading controllers. For each tested parameter pair, the robot traverses a complete 6-meter loop in the testbed at a depth of $0.35$ meters. Tracking ($\delta$) and depth errors for each trial are recorded at $0.6$\,Hz. Table~\ref{tab:performance_metrics} and~\ref{tab:depth_controller_tuning} summarize the observations of the tuning parameters combinations for heading and depth controllers, respectively. We identified that a controller tuned with $K_p = 3.4$ and $K_d = 0.9$ minimized the tracking error, and therefore employed these values as the heading controller gains in all subsequent experiments. Similarly, $K_p = 600$ and $K_d = 50$ are chosen for the depth controller.

\setlength{\extrarowheight}{2pt}
\begin{table}[t]
\centering
\caption{Tracking errors from the line following experiments are compared for various choices of $K_p$ and $K_d$ values of the \textit{heading} controller. $\delta$: mean tracking error (cm), $\sigma$: standard deviation.}%
\vspace{-1mm}
\resizebox{\linewidth}{!}{
\begin{tabular}{c||cccccccccc}
\Xhline{2\arrayrulewidth}
\textbf{$K_p$} & 1 & 2 & 3 & 3 & 3.5 & 3.5 & 3.4 & \textbf{3.4} & 3.5 & 3.4 \\ \hline
\textbf{$K_d$} & 0 & 0 & 0 & 0.5 & 0.5 & 0.7 & 0.7 & \textbf{0.9} & 0.9 & 1.0 \\ \hline
\textbf{$\delta$} & 19.8 & 17.2 & 15.9 & 14.7 & 14.1 & 13.9 & 13.6 & \textbf{13.1}  & 13.4 & 13.3 \\ \hline
\textbf{$\sigma$} & 26.5 & 24.1 & 22.3 & 21.6 & 20.9 & 21.2 & 20.3 & \textbf{19.8} & 21.5 & 20.0 \\
\Xhline{2\arrayrulewidth}
\end{tabular}
}
\label{tab:performance_metrics}
\end{table}

\begin{table}[t]
\centering
\caption{Depth errors from the line following experiments are compared for various choices of $K_p$ and $K_d$ of the depth controller. Here, $\mu$: mean depth error (cm), $\sigma$: standard deviation.}%
\vspace{-1mm}
\resizebox{\linewidth}{!}{
\begin{tabular}{c||cccccccccc}
\Xhline{2\arrayrulewidth}
\textbf{$K_p$} & 500 & 500 & 550 & 600 & 600 & 600 & \textbf{600} & 600 & 650 & 720 \\ \hline
\textbf{$K_d$} & 0 & 10 & 10 & 10 & 20 & 30 & \textbf{50} & 100 & 200 & 300 \\ \hline
\textbf{$\mu$} & 3.09 & 2.98 & 2.47 & 2.39 & 2.06 & 2.61 & \textbf{1.91} & 1.80 & 2.98 & 1.92 \\ \hline
\textbf{$\sigma$} & 2.14 & 1.97 & 1.72 & 1.63 & 1.38 & 1.84 & \textbf{1.38} & 1.70 & 2.02 & 1.44 \\
\Xhline{2\arrayrulewidth}
\end{tabular}
}
\label{tab:depth_controller_tuning}
\vspace{-4mm}
\end{table}

\begin{figure}[t]
    \centering
    \includegraphics[width=\columnwidth]{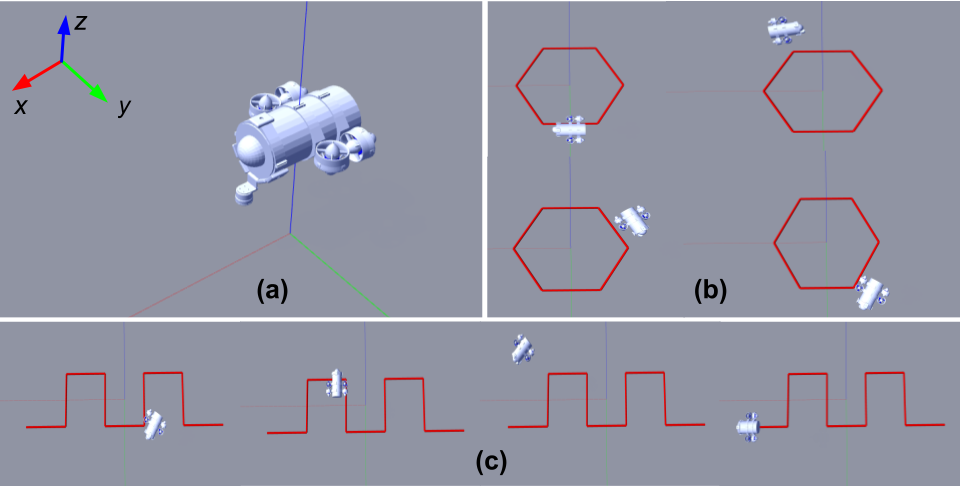}%
    \caption{The digital twin (DT) of CavePI, modeled in ROS, is used for virtual testing in Gazebo underwater environment. (a) An isometric view of the DT in Gazebo; (b,c) Line following experiments in a hexagonal loop and a lawn-mower pattern, respectively.
    }%
    \vspace{-1mm}
    \label{fig:gazebo}
\end{figure}

\begin{figure}[b]
    \centering    \includegraphics[width=0.98\linewidth]{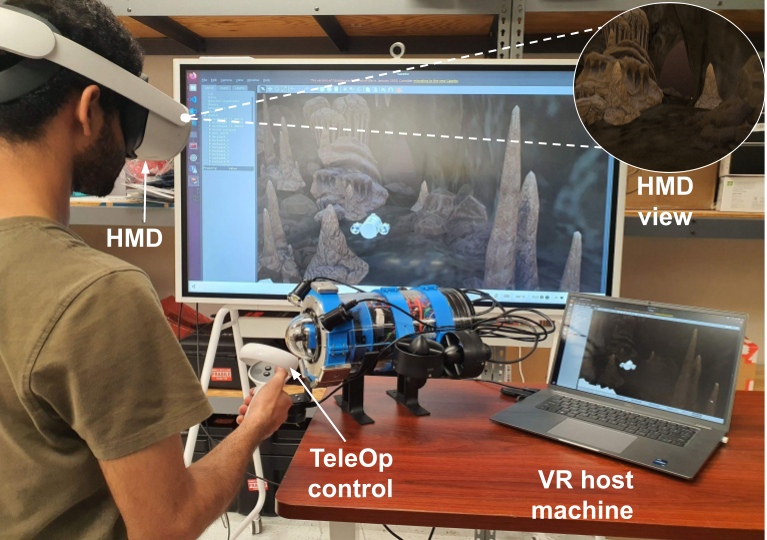}%
    \vspace{-1mm}
    \caption{Interactive demonstration setup for CavePI's virtual missions is shown. A user wears an HMD to experience various camera perspectives while using a handheld controller.
    }
    \label{fig:vr}
\end{figure}

\subsection{ROS Prototyping and Gazebo Simulation}
The digital twin of CavePI is developed using ROS and simulated within Gazebo’s underwater environment to evaluate its line-tracking performance; see Fig.~\ref{fig:gazebo}. As a simplified representation of the physical platform, it enables extensive pre-deployment testing of mission-critical functionalities. 
The digital twin is designed to be positively buoyant, consistent with the actual CavePI, and receives depth-control inputs to maintain a certain depth. Simulating the digital twin under realistic underwater conditions allows for detailed analysis of hydrostatic and hydrodynamic forces, informing component placement for stable buoyancy and achieving near-zero roll and pitch angles when stationary. A complete sensor suite is integrated to confirm optimal camera placement, particularly for the downward-facing camera essential for caveline detection. Additionally, it incorporates a PID controller to simulate CavePI’s line-tracking functionality, serving as a platform for preliminary control algorithm design and validation. Nevertheless, final controller tuning is performed on the physical system through experimental trials due to factors such as nonlinear drag forces, caveline detection dependencies, and design parameter variations between the CAD model and the actual hardware, which cannot be replicated in simulation. We also developed a VR interface for live interactive simulations and immersive experiments. In this interface, users wear a head-mounted display (HMD) to engage in an immersive virtual mission experience; they experience an egocentric view (\ie, robot's perspective) as well as an external global perspective in real-time while the DT model of the robot conducts a mission. Additionally, they can switch from autonomous mode to teleoperation mode and directly control the CavePI in different virtual worlds; see Fig.~\ref{fig:vr}.

%% file: src/06_Field_Trials.tex
\section{Field Experimental Demonstration}

\subsection{Field Trials and Experimental Setups}
The guided navigation capability of CavePI was demonstrated in two distinct real-world environments: (1) shallow riverine areas ($2$\,m - $6$\,m depth) near springs' outlets; and (2) deep natural underwater grottos and caves ($15$\,m - $30$\,m depth).  
The experiments included $15$ open-water trials in spring areas and $10$ trials inside underwater grottos. In the open-water trials, a $20$-meter rope line was laid in both irregular loop patterns and linear configurations along the uneven riverbed. In contrast, cave trials used an actual caveline, which varied in color, texture, and thickness. Each environment posed unique challenges. Strong currents near the spring outlets caused significant drift, particularly when navigating across or against the flow. Additionally, low-light conditions inside underwater caves hindered the accurate detection of the semantic markers.

\begin{figure}[h]
    \centering
    \includegraphics[width=\columnwidth]{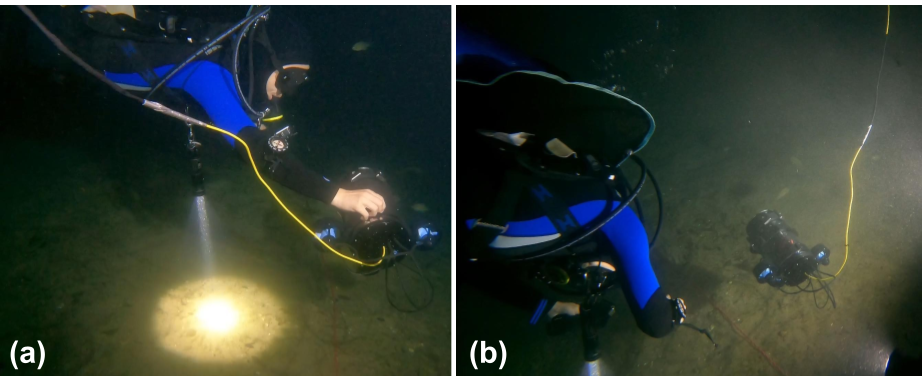}%
    \vspace{-1mm}
    \caption{(a) A support diver is placing the robot on a caveline to initiate tracking; (b) two divers are following the robot once tracking is initiated. Note that the tether is used only for remotely aborting the mission in case of emergency.
    }%
    \label{fig:divers}
\end{figure}

Field deployments were conducted under the supervision of two support divers responsible for initiating and concluding the missions. Divers carried QR-code tags to issue visual commands to CavePI via its front-facing camera. Upon receiving the \texttt{start} command, the robot activated depth-hold mode and began line-following while maintaining the specified depth. A surface station operator monitored each mission remotely. For cave deployments, a tether connected the robot to the surface station to enable remote emergency intervention if necessary (see Fig.~\ref{fig:divers}). These field trials' outcomes highlighted both the strengths and potential areas for improvement in CavePI's perception, planning, and control mechanisms. 

\begin{figure}[h]
\centering
\begin{subfigure}[]{0.98\linewidth}
\includegraphics[width=\linewidth]{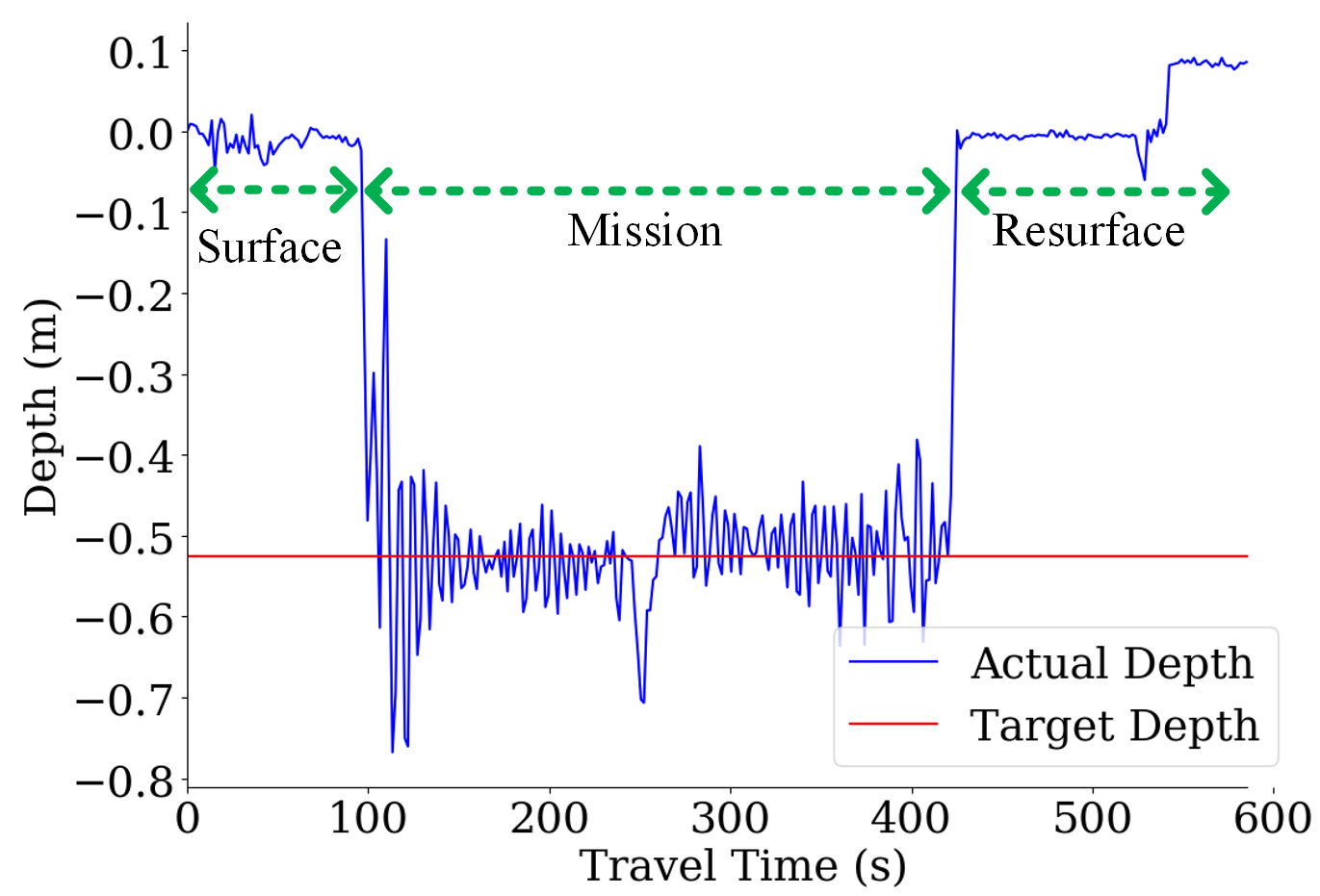}%
\caption{Depth Control Accuracy}
\end{subfigure}
\begin{subfigure}[]{0.98\linewidth}
\includegraphics[width=\linewidth]
{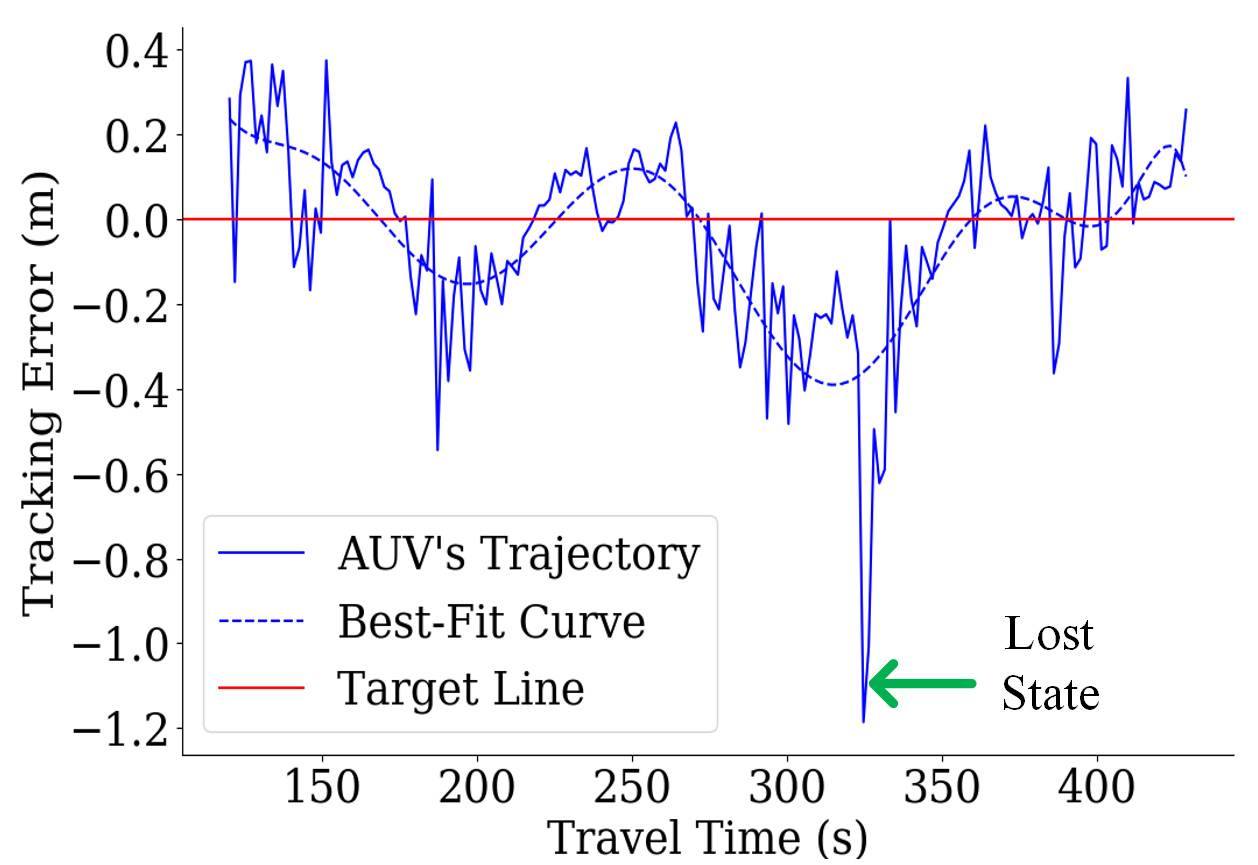}%
\caption{Tracking Accuracy}
\end{subfigure}
\caption{Line-following and depth-holding accuracy are reported for an open-water experiment: (a) the target depth was set to $0.525$\,m; (b) the higher tracking error is caused by strong currents, leading to loss of tracking on one occasion.
}
\label{fig:field_results}
\vspace{-1mm}
\end{figure}

\begin{figure*}[t]
     \centering
     \includegraphics[width=\linewidth]{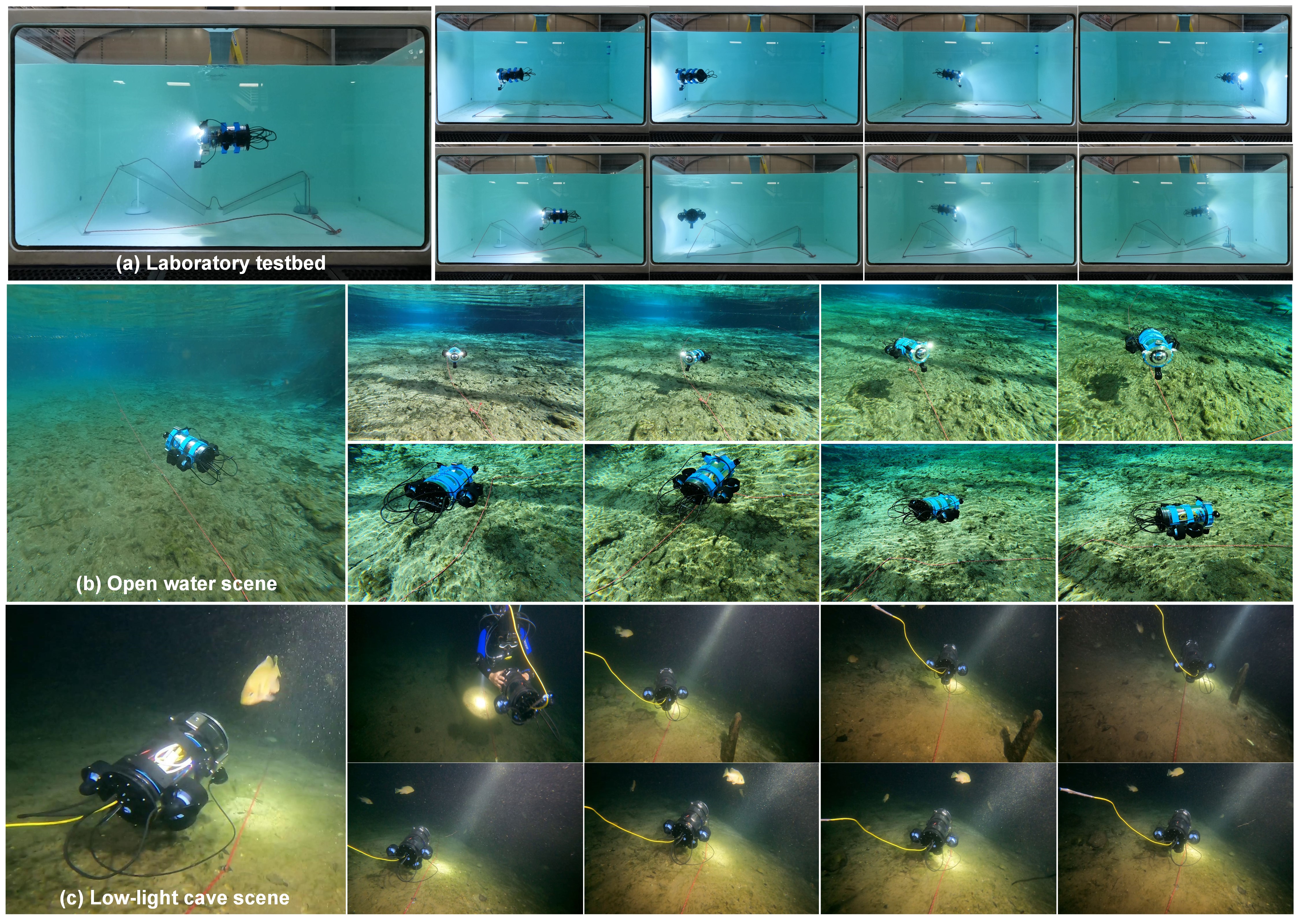}%
     \vspace{-1.5mm}
     \caption{A few snapshots from our field trials for caveline tracking and following experiments with CavePI are shown. The setups include: (a) rectangular loops and slopes in a water tank; (b) irregular shapes in Spring riverine water, and (c) low-light underwater cave scenarios (note that the tether is only for safety and easy recovery). Images are best viewed digitally at $2\times$ zoom; full video demonstration is available online at: \url{https://youtu.be/9BPpB1nu98E}.
     }%
     \vspace{-3mm}
     \label{fig:all_setup}
 \end{figure*}

\subsection{Results and Observations}

\vspace{1mm}
\noindent
\textbf{Laboratory tank tests.} Sec.~\ref{sec_5b} details the setup and evaluation procedures for assessing CavePI's line-following performance in laboratory tests. During these tests, CavePI operated at a depth of $0.35$ meters below the water surface. The depth controller demonstrated robust performance, maintaining the desired depth with a mean error of approximately $2$\,cm. At the beginning of each trial, heading control signals were disabled for $10$\,seconds to allow CavePI to stabilize at the target depth after its descent. Consequently, higher tracking errors were observed during this initial period, as shown earlier in Fig.~\ref{fig:deviation}. Additionally, occasional perception failures were encountered, where the tank edges were misidentified as caveline segments (see Fig.~\ref{fig:perception_failure}\,a), leading to \textit{overshooting}. Despite these transient deviations, CavePI successfully completed multiple loops, demonstrating the robustness of its control strategy.

\vspace{1mm}
\noindent
\textbf{Open-water tests: Spring riverine environments.} These trials were evaluated using the same methodology described in Sec.~\ref{sec_5b}, focusing on both line-tracking and depth-regulation accuracy. Fig.~\ref{fig:field_results} presents a representative result from a 10-minute open-water experiment. As expected, the field results exhibit greater variation compared to the controlled laboratory experiments due to environmental disturbances. Despite these challenges, the depth controller demonstrated robust performance, maintaining a depth within $\pm 10$\,cm without requiring trial-specific parameter tuning.

In contrast, the heading controller proved less resilient under strong currents, resulting in much higher tracking errors. As illustrated in Fig.~\ref{fig:field_results}\,b, the system experienced significant lateral drift, occasionally causing CavePI to lose the caveline. This limitation became particularly pronounced when traveling perpendicular to the current (see Fig.~\ref{fig:tracking_failure}\,b), as there was no active lateral control to counter crossflow. Additionally, the vehicle’s slightly back-heavy design induced a pitch-up motion during upstream travel, as shown in Fig.~\ref{fig:tracking_failure}\,c, further compromising stability and navigation accuracy.

\vspace{1mm}
\noindent
\textbf{Low-light tests: nighttime cave environments.} These trials utilized an earlier (ablation) iteration of CavePI equipped with a three-thruster configuration, in contrast to the newly proposed four-thruster design. The three-thruster setup required precise orientation of the downward thruster to achieve proper heave motion; otherwise, thrust forces could inadvertently induce \textit{roll}. Additionally, the absence of a dedicated roll-control mechanism posed challenges for stabilization. In contrast, the four-thruster configuration incorporates two thrusters for heave motion, enabling dedicated roll control and significantly improving overall stability during navigation.

Despite being equipped with two onboard lights, CavePI faced considerable challenges in low-light environments due to glare and backscattering effects, which impeded accurate semantic perception. Nighttime trials conducted in an underwater cave/grotto system revealed that the lightweight segmentation models struggled to detect the caveline from camera images. During two one-hour dives, support divers reported instances where submerged tree roots, mosses, and other thin structures were mistakenly identified as the caveline. However, once manually repositioned onto the caveline by the divers, CavePI was able to maintain tracking for up to one minute. These findings underscore the need for integrating a more powerful computational platform capable of running robust segmentation models to enhance CavePI’s perception capabilities in challenging environments.

%% file: src/07_Limitation.tex
\section{Limitations \& Challenges}

\subsection{Design Aspects}
CavePI’s onboard components are packed within a single pressure-sealed tube, leaving minimal space for additional hardware. Notably, major housing volume is occupied by the LiPo battery, which can be replaced with a more compact alternative to create space for future add-ons. We also plan to upgrade the perception SBC from the existing Jetson Nano to a Jetson Orin Nano that offers $4\times$ memory with more advanced GPU resources. The redesigned system will also incorporate a magnetic switching mechanism for seamless power control. Outside of the enclosure, the 4 thrusters are arranged to allow control over surge, heave, roll, and yaw motions, however, their adjacent placement induces mutual turbulence and reduces thrust efficiency. To address this, we are investigating alternative thruster placements that will enhance motion dynamics without affecting the robot's buoyancy properties and dynamic stability.

\begin{figure}[t]
    \centering
    \includegraphics[width=\columnwidth]{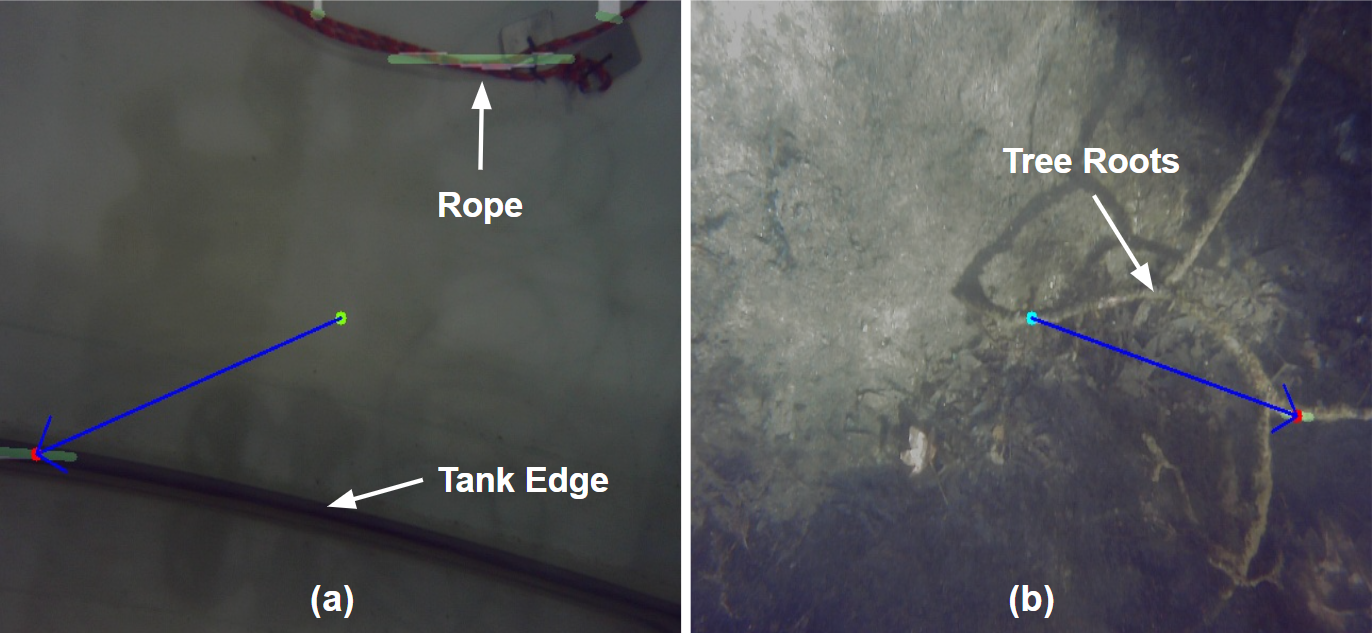}%
    \caption{A few perception failure modes are shown. The down-facing camera falsely detects (a) the edge of the laboratory tank; (b) tree roots as the caveline, resulting in incorrect tracking.
    }%
    \vspace{-3mm}
    \label{fig:perception_failure}
\end{figure}

\subsection{Perception Challenges}
The current onboard sensor suite offers a limited understanding of the surrounding 3D environment. For instance, the Ping2 sonar provides only 1D depth measurements, which we will replace with an advanced $360^\circ$ scanning sonar system for enhanced spatial awareness. The scanning sonar, combined with other state estimation sensors, will map the environment and effectively avoid obstacles during navigation. Moreover, the two cameras currently operate independently for different purposes without synchronization. In the next iteration, we propose incorporating a $45^\circ$ slanted camera and combining all the visual feeds into a mosaic vision. This advanced vision system will offer wider FOV with more peripheral information and improve visual servoing performance. Additionally, the current caveline detection model occasionally misidentifies objects such as submerged tree roots as part of the caveline as shown in Fig.~\ref{fig:perception_failure}\,b, and it sometimes fails to detect the line under low-light conditions. These issues lead to significant tracking inaccuracies. Although more computationally intensive models might address these shortcomings, they are currently infeasible due to hardware limitations. With the proposed design modifications for the next iteration, we plan to adopt a more robust detection model to improve CavePI's tracking performance. Furthermore, hand gesture recognition~\cite{xu2008natural} will be integrated to enable seamless cooperation between divers and CavePI during underwater cave operations. With this advanced sensor setup, we will deploy advanced SLAM algorithms, such as SVIn2~\cite{RahmanIJRR2022} for robust navigation in GPS-denied underwater cave environments. 


\begin{figure}[h]
    \centering
    \includegraphics[width=\columnwidth]{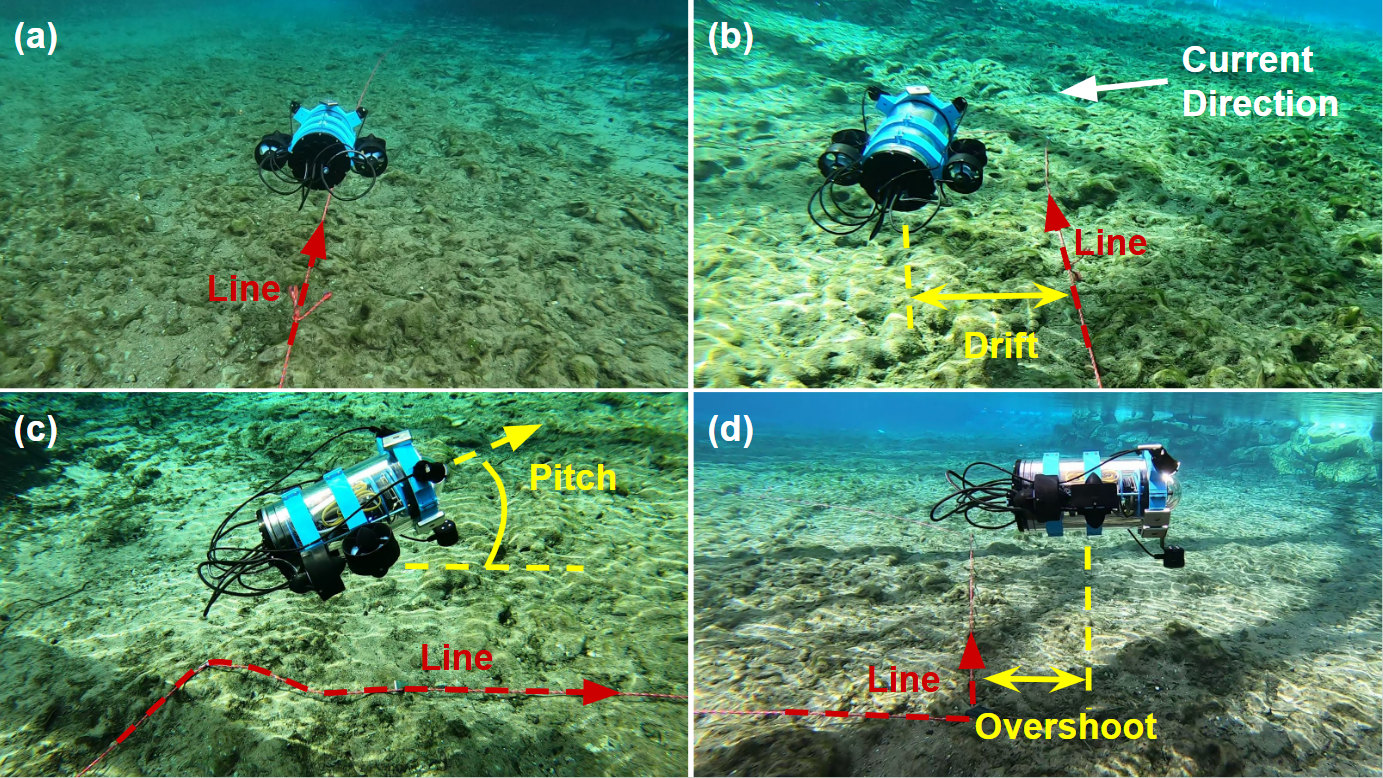}%
    \caption{A few tracking failure modes are shown: (a) CavePI is correctly following the line; (b) lateral drift under strong currents; (c) pitching upward due to currents, attributed to its back-heavy design; (d) overshooting its intended trajectory while moving downstream.
    }%
    \vspace{-1mm}
    \label{fig:tracking_failure}
\end{figure}

\subsection{Smooth 6-DOF Control}
Although CavePI is a 6-DOF AUV capable of maneuvering in a 3D environment, it currently offers active control over four DOFs: surge, heave, roll, and yaw. In future iterations, we intend to reposition its thrusters to enable control over the remaining two DOFs -- pitch and sway -- thereby enhancing the robot’s maneuverability in complex underwater environments. Furthermore, CavePI's autonomous control primarily utilizes a proportional-derivative (PD) controller, which performs effectively under conditions with minimal environmental disturbances. However, this straightforward control strategy becomes unstable in more complex scenarios, such as when water currents are present. In such conditions, lower proportional gains are insufficient for adjusting CavePI’s yaw to align with the cave line, while higher proportional gains cause overshooting in the robot's trajectory during sharp turns (see Fig.~\ref{fig:tracking_failure}\,d), attributed to both perception latency and the limitations of the current control method. Additionally, nonlinear (primarily quadratic) drag forces significantly impact the robot's stability and must be incorporated into the control system design. To overcome these challenges, we are developing a more robust nonlinear adaptive control system to enhance stability. Concurrently, we are optimizing the communication between the perception and control modules to reduce latency and improve overall responsiveness.

%% file: src/08_Review_Discussion.tex


\section{Discussion}
Our investigation identifies four primary challenges impeding effective visual servoing and safe navigation of AUVs within underwater cave environments: (\textbf{1}) accurate scene parsing under noisy sensing conditions, with very little ambient light; (\textbf{2}) robust state estimation in feature-scarce environments lacking GPS or external localization aids; (\textbf{3}) reliable tracking and control mechanisms to maintain visibility of navigation markers during maneuvering; and (\textbf{4}) the inherent logistical complexity and risks involved in field deployments for in-situ experimental validation. These challenges are further exacerbated in our experiments by the compact form factor, low inertia, and energy-efficient design of the CavePI platform, which collectively reduce motion stability in the presence of unpredictable hydrodynamic disturbances, particularly near cave entrance and spring outlets. Furthermore, the performance of our semantic-guided navigation framework hinges on the accuracy of the underlying semantic segmentation model, which is challenged by variations in waterbody types, turbidity, and backscatter commonly encountered in confined and cluttered cave systems. These findings highlight the pressing need for more robust perception and control strategies tailored to the unique operational demands of underwater caves.

To this end, the broader research community stands to benefit from our open documentation of system performance under real-world conditions for both successful outcomes and failure cases. These findings emphasize the critical need to balance perception accuracy with computational efficiency to enhance mission reliability in natural underwater environments -- an aspect often underrepresented in controlled laboratory settings. Another key insight from our field deployments is the influence of external disturbances on caveline-following performance. Our results indicate that effective \textit{sway control} is essential for enabling compact marine robotic systems to maintain stability and trajectory in the presence of strong currents. This observation underscores the need for further research into adaptive control strategies that can dynamically respond to environmental disturbances. Moreover, our integration of visual tags as a mechanism for human-robot interaction offers a scalable and practical alternative to tethered communication for short-range underwater tasks, advancing the potential for more flexible and intuitive operations.

Building upon this foundational system, our ongoing work focuses on advancing toward fully autonomous underwater cave exploration with greater resilience and mission safety. Recent upgrades include the integration of more powerful compute (Nvidia\texttrademark{} Jetson AGX Orin instead of Orin Nano), an adaptive control framework to dynamically compensate for external disturbances, and multi-modal sensor fusion~\cite{talebi2024blueme} to enhance perception robustness and state estimation accuracy. 
Ongoing work is also integrating a more dense segmentation pipeline (using CaveSeg~\cite{abdullah2023caveseg}) for detecting overhead structures and companion divers from sparse pointcloud (by SVIn2~\cite{RahmanIJRR2022}) to improve autonomy. Furthermore, the ability to operate in partially mapped or entirely unknown caves introduces opportunities for innovation in semantic SLAM, risk-aware path planning, and safety-critical control. These enhancements will be particularly valuable in human-robot collaborative missions, where the robot must support shared situational awareness and dynamic decision-making in real-time.

%% file: src/09_Conclusion.tex
\section{Conclusion \& Future Work} \label{sec:conclusion}

Underwater cave exploration is important in hydrology for freshwater resource management, as well as in geological and archaeological studies. Since cave surveys are labor-intensive and hazardous, fully autonomous platforms can help significantly.
This work demonstrates CavePI, a portable and cost-effective AUV tailored for semantic-guided navigation in GPS-denied underwater environments. The platform’s key strength lies in extracting sparse semantic cues from noisy, low-light conditions through a fast and precise segmentation pipeline, enabling intelligent and adaptive navigation. The planning and control systems have been meticulously designed, rigorously tested, and fine-tuned within controlled laboratory settings to ensure robust and reliable performance. Comprehensive field deployments in diverse, challenging underwater environments, along with extensive digital twin simulations, have demonstrated CavePI's reliability, adaptability, and operational efficiency. These results underscore its potential as a versatile and dependable tool for safe, long-term autonomous exploration of underwater caves. Future iterations will optimize system performance and capabilities to support a wider range of applications for exploring any overhead structures (ship hulls, pipelines) without predefined markers. To achieve this, future work will focus on enabling zero-shot semantic understanding and active planning capabilities.

\vspace{1mm}
\section*{Acknowledgment}
\vspace{-1mm}
This research is supported in part by the U.S. National Science Foundation (NSF) grants \#$2330416$, \#$2024741$, and \#$1943205$; and the University of Florida (UF) Research grant \#$132763$. We are thankful to Dr. Nare Karapetyan, Ruo Chen, and David Blow for facilitating our field trials at Ginnie open-water springs and Blue Grotto. Furthermore, we acknowledge the help from Woodville Karst Plain Project (WKPP), El Centro Investigador del Sistema Acuífero de Quintana Roo A.C. (CINDAQ),  Global Underwater Explorers (GUE), Ricardo Constantino, and Project Baseline over the years in providing access to challenging underwater caves and mentoring us in underwater cave mapping. The authors are also grateful for equipment support by Halcyon Dive Systems, Teledyne FLIR LLC, and KELDAN GmbH lights.